\documentclass{article}

\usepackage{microtype}
\usepackage{graphicx}
\usepackage{subfigure}
\usepackage{booktabs} 
\usepackage{multirow}
\usepackage{enumitem}
\usepackage{float}
\restylefloat{table}
\usepackage{nicematrix}
\usepackage{float}

\usepackage{hyperref}

\usepackage[accepted]{icml2024}

\usepackage{amsmath}
\usepackage{amssymb}
\usepackage{mathtools}
\usepackage{amsthm}

\usepackage[capitalize,noabbrev]{cleveref}

\theoremstyle{plain}

\theoremstyle{definition}

\theoremstyle{remark}

\usepackage[textsize=tiny]{todonotes}

\icmltitlerunning{Semantic Space Informed Prompt Learning with LLM for Time Series Forecasting}

\begin{document}

\twocolumn[
\icmltitle{$\textbf{S}^2$IP-LLM: Semantic Space Informed Prompt Learning with LLM for \\ Time Series Forecasting}




\begin{icmlauthorlist}
\icmlauthor{Zijie Pan}{uconn}
\icmlauthor{Yushan Jiang}{uconn}
\icmlauthor{Sahil Garg}{ms}
\icmlauthor{Anderson Schneider}{ms}
\icmlauthor{Yuriy Nevmyvaka}{ms}
\icmlauthor{Dongjin Song}{uconn}
\end{icmlauthorlist}
\icmlaffiliation{uconn}{School of Computing, University of Connecticut, Storrs, USA}
\icmlaffiliation{ms}{Department of Machine Learning Research, Morgan Stanley, New York, USA}

\icmlcorrespondingauthor{Yuriy Nevmyvaka}{yuriy.nevmyvaka@ morganstanley.com}
\icmlcorrespondingauthor{Dongjin Song}{dongjin.song@uconn.edu}

\icmlkeywords{Machine Learning, ICML}

\vskip 0.3in
]




\printAffiliationsAndNotice{}

\begin{abstract}

Recently, there has been a growing interest in leveraging pre-trained large language models (LLMs) for various time series applications. However, the semantic space of LLMs, established through the pre-training, is still underexplored and may help yield more distinctive and informative representations to facilitate time series forecasting. To this end, we propose Semantic Space Informed Prompt learning with LLM ($S^2$IP-LLM) to \emph{align the pre-trained semantic space with time series embedding space} and perform time series forecasting based on learned prompts from the joint space. We first design a tokenization module tailored for cross-modality alignment, which explicitly concatenates patches of decomposed time series components to create embeddings that effectively encode the temporal dynamics. Next, we leverage the pre-trained word token embeddings to derive semantic anchors and align selected anchors with time series embeddings by maximizing the cosine similarity in the joint space. This way, $S^2$IP-LLM can retrieve relevant semantic anchors as prompts to provide strong indicators (context) for time series that exhibit different temporal dynamics. With thorough empirical studies on multiple benchmark datasets, we demonstrate that the proposed $S^2$IP-LLM can achieve superior forecasting performance over state-of-the-art baselines. Furthermore, our ablation studies and visualizations verify the necessity of prompt learning informed by semantic space.

\end{abstract}

\section{Introduction}

Over the past few years, pre-trained large language models (LLMs) such as GPT-4~\citep{Achiam2023GPT4TR} and LLaMA~\citep{touvron2023llama1,touvron2023llama2} not only achieved great success across a diverse
range of natural language processing (NLP) tasks, \textit{i.e.}, generate coherent and contextually relevant text, answer questions, and translate sentences between multiple languages, but also exhibited tremendous potential in tackling applications of more complex or structured domains, such as code generation,  healthcare, finance, and autonomous systems, \textit{etc}~\citep{singhal2022large,cui2024survey,li2023large}. As time series analysis is becoming increasingly important for strategic planning and operational efficiency in various real-world applications, \textit{e.g.}, energy load management, traffic forecasting, weather forecasting, health risk analysis, \textit{etc}~\citep{friedman1962interpolation,courty1999timing,bose2017probabilistic,gao2020robusttad,li2022demand,liu2023sadi,dimri2020time}, a natural question to ask is \emph{whether we should train a general purpose foundation model from scratch, or fine-tune pre-trained LLMs to perform time series forecasting?}

Recently, significant efforts have been made to build foundation models for general-purpose time series analysis~\citep{wu2023timesnet,garza2023timegpt,rasul2023lag}. TimesNet~\citep{wu2023timesnet} uses TimesBlock as a task-general backbone to capture multi-periodicity and extract complex intraperiod- and interperiod-variations via transformed 2D tensors. TimeGPT-1 describes a general pre-trained model for time series forecasting ~\cite{garza2023timegpt}. These approaches, however, are hindered by two main challenges. First, time series data can be acquired in various formats, such as univariate or multivariate, often in large volumes, and from different domains, like healthcare, finance, traffic, environmental sciences, \textit{etc}. This escalates the complexity of model training and poses challenges in handling different scenarios. Second, time series data, in practice, often exhibit non-stationary characteristics, resulting in the underlying statistical properties, such as means, variances, and auto-correlations shifting during collection. This could also result in concept drift, where the statistical properties of target variables change over time. These realities present significant challenges for large models to be adapted and retrained effectively.

On the other hand, LLMs trained on extensive and diverse text corpora can serve as a foundational knowledge base that can be applied to a variety of downstream tasks with minimal task-specific prompt learning or fine-tuning. Inspired by this, there has been a growing interest in leveraging existing LLMs to facilitate time series analysis. For instance, \citet{zhou2023onefitsall} utilizes a frozen pre-trained language model to attain state-of-the-art or equivalent performance. \citet{jin2023time} develop time-LLM to reprogram the input time series via text prototype representations by incorporating the embeddings of the dataset's text descriptions as context information. In real-world applications, however, dataset description information may not always be available or informative. In addition, the patching operation (\textit{i.e.}, tokenization), which splits a long time series sequence into overlapping segments over instance normalized time series input, may have limited expressibility as it could fail to capture the subtle variations of different components in time series.

\begin{figure}[t]
\begin{center}
\centerline{\includegraphics[width=\columnwidth]{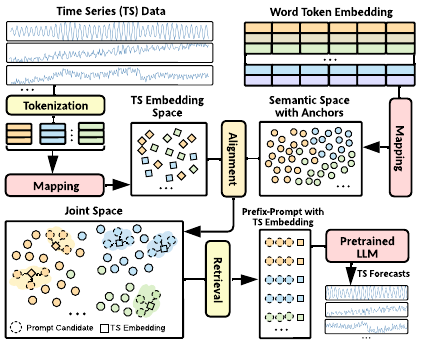}}
\caption{The demonstration of semantic space informed prompting in $S^{2}$\text{IP-LLM}. The input time series is decomposed and mapped to obtain time series (TS) embedding. Next, the TS embedding is aligned with semantic anchors derived from the pre-trained word token embedding. Finally, top-$K$ similar semantic anchors are retrieved and used as prefix-prompts with TS embedding.}
\label{motivation}
\end{center}
\end{figure}

In this paper, we argue that the semantic space in the form of word token embeddings (based on pre-trained LLMs) can already offer a more distinctive and informative representation space~\citep{ethayarajh2019contextual} to help align time series embeddings. Based on this, we develop Semantic Space Informed Prompt with LLM ($S^2$IP-LLM) for time series forecasting. Specifically, as shown in Figure~\ref{motivation}, we first design a tokenization module tailored to semantic space alignment, which explicitly concatenates patches of decomposed time series components (\textit{i.e.}, trend, seasonality, and residual) to create an embedding that effectively encodes the temporal dynamics more expressively. Next, we map the pre-trained word embeddings to obtain semantic anchors and align selected anchors with time series embeddings by maximizing the cosine similarity in the joint space. 
In this way, $S^2$IP-LLM can retrieve relevant semantic anchors as prefix-prompts to provide strong indicators (context) for time series embeddings that exhibit different temporal dynamics. Our experiments over several standard benchmark datasets demonstrate that $S^2$IP-LLM can achieve superior forecasting performance over state-of-the-art baselines. Moreover, our ablation studies and visualizations also verify the necessity of prompt learning in the joint space.

To summarize, our contributions include:
\begin{itemize}
\itemsep0em 
\parskip0em
    \item We design a specialized tokenization module
    that concatenates patches of decomposed time series components to provide more expressive local contexts and facilitate semantic space informed prompting.
    \item We leverage semantic anchors derived from pre-trained word token embeddings (semantic space) to align time series embeddings and learn a distinctive and informative joint space. Moreover, aligned semantic anchors are used as prompt indicators (contexts) to enhance the representation of time series.
    \item Our experiments and analysis on multiple benchmark datasets demonstrate the superiority of $S^2$IP-LLM over state of the art and the necessity of prompt learning informed by semantic space.
  
\end{itemize}

\section{Related Work}

\subsection{Time Series Forecasting}

In recent years, a variety of statistical and machine learning methods have been developed for time series analysis, \textit{e.g.}, ARIMA~\citep{Anderson1976TimeSeries2E}, Prophet~\citep{taylor2018forecasting}, \textit{etc}. More recently, different types of deep neural networks have been applied for time series analysis. For instance, recurrent neural network (RNN) based models have been developed to capture auto-regressive temporal dynamics~\citep{qin2017dual,li2017diffusion,lai2018modeling,gu2021efficiently}. Graph neural networks (GNN) based methods are leveraged to capture variable dependencies among different time series~\citep{cao2020spectral,wu2020connecting,shang2021discrete,pan2024structural}. Transformer based models leverage the self-attention mechanisms tailored for time series to better capture the temporal dynamics, variable dependencies, or both~\cite{woo2022etsformer,zhou2021informer,wu2021autoformer,zhou2022fedformer,liu2023itransformer}. More recently, MLP-based models~\citep{challu2023nhits,zeng2023transformers} and convolution-based models~\cite{wu2023timesnet} have achieved state-of-the-art performance on par with Transformers, but with much simpler designs. \textcolor{black}{Nevertheless, while these deep forecasters perform well on specific datasets, they lack the flexibility and generalizability to adapt to real-world time series data from different domains.}

\subsection{Pre-trained Large Model for Time Series Analysis}

Recent advancements in natural language processing (NLP) and computer vision (CV) demonstrate that pre-trained models can effectively adapt to a range of downstream tasks through fine-tuning~\citep{bao2021beit,he2022masked,brown2020language,devlin2018bert}. Inspired by this, several different pre-trained models have been developed for time series based on either supervised~\citep{fawaz2018transfer} or self-supervised learning~\citep{zhang2022self,deldari2022beyond}. During the training stage, models can learn robust representations from a variety of input time series data. Then, these models can be fine-tuned for downstream tasks of similar domains to further enhance their performance~\citep{tang2022domain}. With the emergence and success of Large Language Models (LLMs), including T5~\citep{raffel2020exploring}, GPT-based models~\citep{radford2018improving,radford2019language,brown2020language,ouyang2022training}, and LLaMA~\citep{touvron2023llama}, which have showcase their robust pattern recognition and reasoning abilities over complex sequences of tokens, there is a trend to explore how to effectively transfer knowledge from these powerful pre-trained LLM models to time series domain~\citep{jiang2024empowering}. One line of research focuses on leveraging the pre-trained LLMs as zero-shot learners. For instance,~\citet{xue2022promptcast} and~\citet{gruver2023llmtime} directly convert time series data to corresponding text sequence inputs and achieve encouraging results for time series forecasting. Another line of research~\cite {zhou2023onefitsall,chang2023llm4ts} involves tokenizing the input time series data into overlapping patches and strategically leveraging or fine-tuning LLMs for time series analysis. Following this paradigm, TEST~\citep{sun2023test} and Time-LLM~\citep{jin2023time} reprogram time series data with text prototype embedding and incorporate textual prompts for time series analysis. TEMPO~\citep{cao2023tempo} incorporates the decomposition of time series and retrieval-based prompt design for non-stationary time series data. Different from those methods, we explicitly leverage semantic anchors derived from pre-trained word token embeddings (semantic space) to align time series embeddings and develop a simple yet effective prompt mechanism to inform LLM for forecasting tasks.

\section{Methodology}

\begin{figure*}[htbp] 
  \centering
  \includegraphics[width=\textwidth]{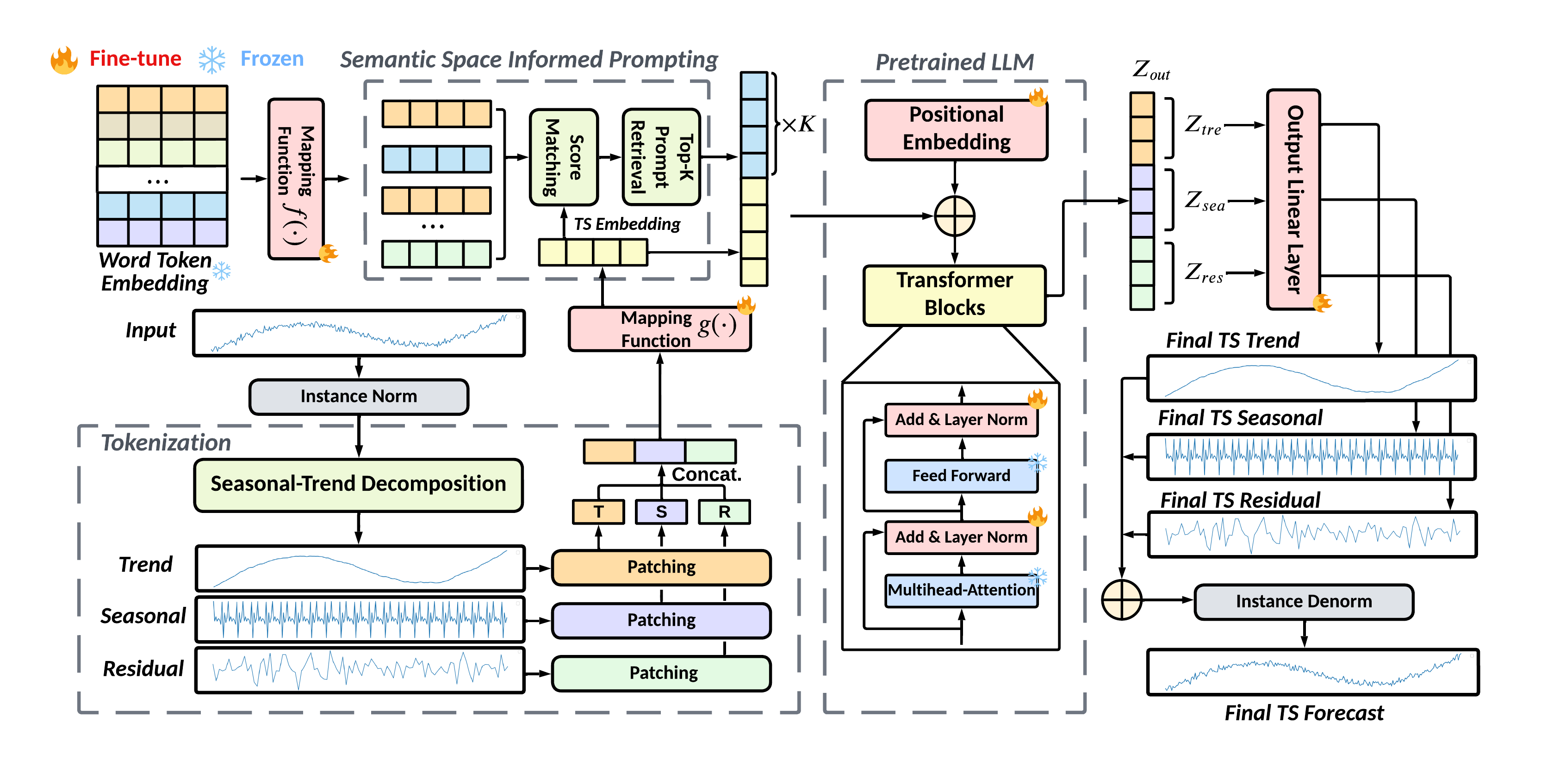} %
  \caption{\textcolor{black}{The model architecture of $S^{2}\text{IP-LLM}$. The input time series is normalized, decomposed, patched individually, and concatenated to represent the context of time series (TS). Semantic space informed prompting performs alignment between the contextual TS embeddings and the semantic anchors extracted from pre-trained word embeddings, and retrieves the most similar $K$ ones as prefix-prompts. The decomposed TS representations from pre-trained LLM are linearly projected and combined as the TS forecast.}}
  \label{pipeline} 
  
\end{figure*}

\textbf{Overview}: $S^{2}$IP-LLM consists of three key components as shown in Figure \ref{pipeline}. Given the input time series, we first tokenize it and obtain the time series (TS) embedding based on \textcolor{black}{time series decomposition and patching. }
Next, we will align the TS embedding with semantic anchors derived from the pre-trained word token embedding. Finally, top-$K$ similar semantic anchors will be retrieved to serve as prefix-prompts for the TS embedding and the concatenated vector will be leveraged as the query for pre-trained LLMs. 

In this paper, GPT-2 is used as the backbone. During the training stage, we not only learn the mapping functions of input and output but also fine-tune the positional embedding and layer norm block of GPT-2. 

\subsection{Problem Statement}

We first formalize the time series forecasting problem. Let $X \in \mathbb{R}^{N \times T}$ denote the time series data containing $N$ variables and $T$ time steps, where $X_{:,t}\in \mathbb{R}^{N \times 1}$ denotes $t$-th time step across all variables and $X_{i,:}\in \mathbb{R}^{1 \times T}$ denotes $i$-th variable. Given a historical $\tau$-step window of time series, we aim to learn a forecasting module $\mathcal{F \left( \cdot \right)}$ that will predict the next $\tau'$ time steps based on the input window. Mathematically, at a starting time step $t$, the corresponding forecast is given by $\hat{Y} = \hat{X}_{:,t:t+\tau'-1} = \mathcal{F}(X_{:,t-\tau:t-1})$.

\subsection{Time Series Tokenization}

\textcolor{black}{In real-world applications, non-stationary data is prevalent. To tackle this problem, we first apply the reversible instance normalization \citep{kim2021reversible} on time series input such that the data has zero mean and unit standard deviation to mitigate the distribution shift in time series. Specifically, given the $i$-th time series input at time step $t$, \textit{i.e.}, $X_{i,t}$, the transformed value $X_{i, t}^{\prime}$ can be given by:
\begin{equation}  
 X_{i, t}^{\prime}=\gamma_T\left(X_{i, t}-\frac{\mathbb{E}_t\left[X_{i, t}\right]}{\sqrt{\operatorname{Var}\left[X_{i, t}\right]+\epsilon_T}}\right)+\beta_T
\end{equation}
where $\mathbb{E}_t\left[X_{i, t}\right]$ and $\operatorname{Var}\left[X_{i, t}\right]$ are the instance-specific mean and variance, respectively. $\gamma_T$ and $\beta_T$ are trainable parameters. Next, we adopt an additive seasonal-trend decomposition method to decompose normalized time series into long-term trend, seasonal, and residual components. The additive seasonal-trend decomposition is given by $X_{i, t}^{\prime}=X_{i, t}^{\text {tre}\prime}+X_{i, t}^{\text {sea}\prime}+X_{i, t}^{\text {res}\prime}$, where $\textit{tre},\textit{sea},\textit{res}$ denotes the long-term trend, seasonal, and residual component, respectively. There are several options for additive seasonal-trend decomposition. One option is the classical additive seasonal-trend decomposition that first obtains long-term trend components using moving averages. Then, the seasonal component is estimated by averaging the detrended time series with pre-defined season parameters. Finally, the residual component is obtained by subtracting the estimated trend and seasonal components from the normalized time series. Another option is the Seasonal-Trend decomposition using Loess (STL)~\citep{cleveland1990stl}. The choice of decomposition method will based on validation results.}



Next, we follow~\citep{Yuqietal-2023-PatchTST} to encode temporal information and local contexts of input time series by aggregating consecutive time steps into overlapped patched tokens. Take the trend component as an example, the normalized component series, $X_{i, t-\tau: t-1}^{\text{tre }\prime} \in \mathbb{R}^{1 \times \tau}$ is converted to patched token representation $P^{\text{tre }}_{i,t-\tau:t-1} \in \mathbb{R}^{N_P \times L_P}$, in which $L_P$ is the patch length, $N_P=\left\lfloor\frac{\left(\tau-L_P\right)}{S}\right\rfloor+2$ is the number of patches and $S$ is the horizontal sliding stride. We apply patching to each variable component over the temporal dimension and then concatenate the tokens of these components into a single meta-token, $P_{i,t-\tau:t-1} = \left[P^{\text{tre }}_{i,t-\tau:t-1}, P^{\text{sea }}_{i,t-\tau:t-1}, P^{\text{res }}_{i,t-\tau:t-1}\right]  \in \mathbb{R}^{N_P \times 3L_P}$. We then feed the meta-token into a projection layer $g(\cdot)$ to get the time series embedding $\mathcal{P}_{i,t-\tau:t-1}=g\left(P_{i,t-\tau:t-1}\right) \in \mathbb{R}^{N_P \times D}$, where $D$ is the embedding size for the pre-trained LLMs.

\subsection{Semantic Space Informed Prompting}

Prompting has emerged as an effective technique in various applications, enabling LLMs to utilize task-specific information to achieve enhanced reasoning capabilities~\citep{yin2023survey}. Existing works primarily focus on employing template-based and fixed prompts for pre-trained LLMs in time series analysis ~\citep{xue2022promptcast,jin2023time}. While these methods are intuitive, straightforward, and yield satisfactory results, their rigid prompt contexts are in line with linguistic semantics. However, time series representation inherently lacks human semantics and is more closely tied to sequence patterns in the form of temporal dynamics. Conversely, Lester et al.~(\citeyear{lester2021power}) demonstrate the effectiveness of soft prompts in enabling LLMs to comprehend inputs more effectively. In the realm of time series analysis with LLMs, recent works \citep{sun2023test,cao2023tempo} start to consider soft prompts as task-specific, randomly initialized, trainable vectors that learn from the supervised loss between LLM’s output and the ground truth. However, the semantic space of LLMs, established through the pre-training, is still underexplored and may help yield more distinctive and informative representations for time series data. Based on this intuition, we introduce a prompting mechanism informed by the pre-trained semantic space. Specifically, the pre-trained semantic word token embeddings, represented as $\mathbf{E} \in \mathbb{R}^{V \times D}$ where $V$ is the vocabulary size, are inevitably large and dense. For example, the vocabulary size of GPT-2 ~\citep{radford2019language} reaches 50,257 and may raise computational deficiency. Instead of directly using the semantic word token embedding, we derive a small set of semantic anchors $\mathbf{E}'$ in the hidden space using a generic mapping function $f(\cdot)$ on $\mathbf{E}$, which is denoted as $\mathbf{E}' = f\left( \mathbf{E} \right) \in \mathbb{R}^{V' \times D}$, where $V'$ is the reduced number of semantic anchors and  $V^{\prime} \ll V$. 
To properly retrieve relevant semantic anchors to enhance the time series embedding $\mathcal{P}_{i,t-\tau:t-1}$, we align the semantic anchors and time series embedding based on a score-matching function $\gamma(\cdot)$. In this paper, we implement the score-matching function based on cosine similarity
\begin{equation}
 \gamma\left(\mathcal{P}_{i,t-\tau:t-1}, \boldsymbol{e}^{\prime}_m \right) = \frac{\mathcal{P}_{i,t-\tau:t-1} \cdot \boldsymbol{e}^{\prime}_m }{\left\|\mathcal{P}_{i,t-\tau:t-1}\right\| \left\|\boldsymbol{e}^{\prime}_m\right\| },
\end{equation} 
where $\boldsymbol{e}^{\prime}_m \in \mathbf{E}'$. We select top-$K$ relevant semantic anchors based on the similarity scores and utilize them as prefix-prompt to enhance the input time series embedding, \textit{i.e.}, 

\begin{equation}
\begin{aligned}
\boldsymbol{Z}_{i,t-\tau:t-1}&=\left[\boldsymbol{e}^{\prime}_1 ; \cdots ; \boldsymbol{e}^{\prime}_{\textrm{$K$}} ; \mathcal{P}_{i,t-\tau:t-1} \right] 
\\ 
&= \left[\boldsymbol{e}^{\prime}_{\textrm{top-$K$}} ; \mathcal{P}_{i,t-\tau:t-1} \right]
\end{aligned}
\end{equation}

which will serve as the input for the pre-trained LLMs.

\subsection{Optimization Objective}

We can obtain the output embedding $\boldsymbol{Z_{\text{out }}}$ after the forward path of the prompt enhanced time series embedding through LLMs. We will flatten it and use a linear mapping to project the representation to the forecasting horizon $\boldsymbol{Y_{\text{out }}}$. The overall forecasting should also be the additive combination of the individual component predictions due to the decomposition step. We further split and express $\boldsymbol{Y_{\text{out }}}$ into a concatenation form $\boldsymbol{Y_{\text{out }}} = \left[Y^{\text{tre }}_{\text{out }}, Y^{\text{sea }}_{\text{out }}, Y^{\text{res }}_{\text{out }}\right]$ and obtain the forecasting results as 
$\hat{Y} = Y^{\text{tre }}_{\text{out }} + Y^{\text{sea }}_{\text{out }} + Y^{\text{res }}_{\text{out }}$.
At every training iteration, the overall training objective is:
\begin{equation}
\min \mathcal{L} \left( \hat{Y}, X_{:,t:t+\tau'-1}\right) - \lambda\sum \gamma\left(\mathcal{P}_{i,t-\tau:t-1}, \boldsymbol{e}^{\prime}_{\textrm{top-$K$}} \right),
\end{equation}
where the first term is the forecasting loss in the form of mean squared error (MSE), and the second term is a score-matching function to align selected semantic anchors with the time series embedding obtained via decomposition and patching. In this way, we could obtain a more informative space to facilitate the underlying forecasting task. $\lambda\geq 0$ is a hyper-parameter to trade-off the alignment.

\subsection{Backbone and Fine-tuning Strategy}

In this paper, we employ GPT-2 ~\citep{radford2019language} as our pre-trained large language model (LLM) backbone. We choose to keep a significant portion of the parameters frozen, especially those parameters related to the multi-headed attention and the feed-forward networks within the Transformer blocks. This strategy can not only reduce the computational burden but also align with existing literature ~\citep{lu2022frozen,houlsby2019parameter,zhou2023onefitsall}. They suggest that maintaining most of the parameters in their non-trainable state can achieve better outcomes compared to completely retraining LLMs. For GPT-2, we only fine-tune the positional embedding layer and the layer-normalization layers.

\section{Experiments}

In our experiments, we compare the proposed $S^2$IP-LLM against a variety of baselines on 11 public datasets. We validate the effectiveness of $S^2$IP-LLM over different time series tasks, including long-term forecasting (Section 4.1), short-term forecasting (Section 4.2), and few-shot forecasting (Section 4.3). We also provide the ablation studies and parameter sensitivity analysis in Section 4.4. Finally, we visualize the prompt enhanced time series embeddings to qualitatively assess the effectiveness of $S^2$IP-LLM. We follow the experimental configurations~\citep{wu2023timesnet} for all baselines using the unified pipeline.\footnote{https://github.com/thuml/Time-Series-Library}

\textbf{Baselines.} The baselines include a set of Transformer-based methods, \textit{i.e.}, iTransformer ~\citep{liu2023itransformer}, PatchTST~\citep{Yuqietal-2023-PatchTST}, FEDformer~\citep{zhou2022fedformer}, Autoformer~\citep{wu2021autoformer}, Non-Stationary Transformer~\citep{liu2022non}, ETSformer~\citep{woo2022etsformer} and Informer~\citep{zhou2021informer}. We also select a set of non-transformer based techniques, \textit{i.e.}, DLinear~\citep{zeng2023transformers}, TimesNet~\citep{wu2023timesnet}, and LightTS~\citep{zhang2022less} for comparison. Finally, two approaches based on LLMs, \textit{i.e.}, OFA~\citep{zhou2023onefitsall} and Time-LLM~\citep{jin2023time}\footnote{\textcolor{black}{We reproduced results through public available code:  https://github.com/KimMeen/Time-LLM. For Time-LLM variants, `L' denotes the LLaMA backbone, and `G' refers to the GPT-2 backbone.}}.

\begin{table*}[!h]

\caption{Long-term forecasting results for \{96, 192, 336, 720\} horizons. Lower values indicate better performance. \textcolor{black}{For Time-LLM variants, `L' denotes the LLaMA backbone~\citep{touvron2023llama}, and `G' refers to the GPT-2 backbone~\citep{radford2019language}.} More results are in Appendix~\ref{sec:appendix:long-term}, Table~\ref{tab:long-term-forecast-transformer}}

\setlength{\tabcolsep}{3.8pt}
\footnotesize
\centering
\begin{tabular}{cc|cc|cc|cc|cc|cc|cc|cc}
\hline\hline
\multicolumn{2}{c|}{Methods} &
  \multicolumn{2}{c|}{\textbf{$\mathbf{S^2}$IP-LLM}} &
  \multicolumn{2}{c|}{\textbf{Time-LLM(L)}}&
  \multicolumn{2}{c|}{\textbf{Time-LLM(G)}}&
  \multicolumn{2}{c|}{\textbf{OFA}} &
  \multicolumn{2}{c|}{\textbf{iTransformer}} &
  \multicolumn{2}{c|}{\textbf{Dlinear}} &
  \multicolumn{2}{c}{\textbf{PatchTST}} 
\\ \hline

\multicolumn{2}{c|}{Datasets$\backslash$Horizon} &
  \multicolumn{1}{c}{MSE} &
  \multicolumn{1}{c|}{MAE} &
  \multicolumn{1}{c}{MSE} &
  \multicolumn{1}{c|}{MAE} &
  \multicolumn{1}{c}{MSE} &
  \multicolumn{1}{c|}{MAE} &
  \multicolumn{1}{c}{MSE} &
  \multicolumn{1}{c|}{MAE} &
  \multicolumn{1}{c}{MSE} &
  \multicolumn{1}{c|}{MAE} &
  \multicolumn{1}{c}{MSE} &
  \multicolumn{1}{c|}{MAE} &
  \multicolumn{1}{c}{MSE} &
  \multicolumn{1}{c}{MAE}
  \\ 
  \hline
\multicolumn{1}{l|}{\multirow{5}{*}{\textbf{Weather}}} 
& 96 & \textbf{0.145} & \textbf{0.195} & \underline{0.148} & \underline{0.197} & 0.158 & \multicolumn{1}{c|}{\textcolor{black}{0.210}} & 0.162 & 0.212 & 0.253 & 0.304 & 0.176 & 0.237 & 0.149 & 0.198 \\
\multicolumn{1}{l|}{}                       
& 192 & \textbf{\textcolor{black}{0.190}} & \textbf{\textcolor{black}{0.235}} & \underline{\textcolor{black}{0.194}} & \multicolumn{1}{c|}{\textcolor{black}{0.246}} &  \textcolor{black}{0.197} & \multicolumn{1}{c|}{\textcolor{black}{0.245}} & 0.204 & 0.248 & 0.280 & 0.319 & 0.220 & 0.282 & \underline{0.194} & \underline{0.241}  \\
\multicolumn{1}{l|}{}                       
& 336 & \textbf{\textcolor{black}{0.243}} & \textbf{\textcolor{black}{0.280}} & \textcolor{black}{0.248} & \multicolumn{1}{c|}{\textcolor{black}{0.285}} & \textcolor{black}{0.248} & \multicolumn{1}{c|}{\textcolor{black}{0.285}} & 0.254 & 0.286 & 0.321 & 0.344 & 0.265 & 0.319 & \underline{0.245} & \underline{0.282} \\ 
\multicolumn{1}{l|}{}                      
& 720 & \textbf{\textcolor{black}{0.312}} & \textbf{\textcolor{black}{0.326}} & \textcolor{black}{0.317} & \underline{\textcolor{black}{0.332}} & \textcolor{black}{0.319} & \multicolumn{1}{c|}{\textcolor{black}{0.334}} & 0.326 & 0.337 & 0.364 & 0.374 & 0.333 & 0.362 & \underline{0.314} & 0.334  \\
\multicolumn{1}{l|}{}                       
& Avg. & \textbf{\textcolor{black}{0.222}} & \textbf{\textcolor{black}{0.259}} & \textcolor{black}{0.226} & \multicolumn{1}{c|}{\textcolor{black}{0.265}} & \textcolor{black}{0.230} & \multicolumn{1}{c|}{\textcolor{black}{0.268}} & 0.237 & 0.270 & 0.304 & 0.335 &0.248 & 0.300 & \underline{0.225} & \underline{0.264}   \\

\hline

\multicolumn{1}{l|}{\multirow{5}{*}{\textbf{Electricity}}} 
& 96  & \underline{\textcolor{black}{0.135}} & \underline{\textcolor{black}{0.230}} & \textcolor{black}{0.140} & \multicolumn{1}{c|}{\textcolor{black}{0.246}} &\textcolor{black}{0.137} & \multicolumn{1}{c|}{\textcolor{black}{0.237}} & 0.139 & 0.238 & 0.147 & 0.248 & 0.140 & 0.237 & \textbf{0.129} & \textbf{0.222}   \\
\multicolumn{1}{l|}{}                       
& 192 & \textbf{\textcolor{black}{0.149}} & \underline{\textcolor{black}{0.247}} & \textcolor{black}{0.155} & \multicolumn{1}{c|}{\textcolor{black}{0.253}} & \underline{\textcolor{black}{0.150}} & \multicolumn{1}{c|}{\textcolor{black}{0.249}} & 0.153 & 0.251 & 0.165 & 0.267 & 0.153 & 0.249 & 0.157 & \textbf{0.240}   \\
\multicolumn{1}{l|}{}                       
& 336 & \underline{\textcolor{black}{0.167}} & \underline{\textcolor{black}{0.266}} & \textcolor{black}{0.175} & \multicolumn{1}{c|}{\textcolor{black}{0.279}} & \textcolor{black}{0.168} & \underline{\textcolor{black}{0.266}} & 0.169 & \underline{0.266} & 0.178 & 0.279 & 0.169 & 0.267 & \textbf{0.163} & \textbf{0.259}   \\ 
\multicolumn{1}{l|}{}                      
& 720 & \underline{\textcolor{black}{0.200}} & \textbf{\textcolor{black}{0.287}} & \textcolor{black}{0.204} & \multicolumn{1}{c|}{\textcolor{black}{0.305}} & \textcolor{black}{0.203} & \multicolumn{1}{c|}{\textcolor{black}{0.293}} & 0.206 & 0.297 & 0.322 & 0.398 & 0.203 & 0.301 & \textbf{0.197} & \underline{0.290} \\
\multicolumn{1}{l|}{}                       
& Avg. &  \textbf{\textcolor{black}{0.161}} & \underline{\textcolor{black}{0.257}} & \textcolor{black}{0.168} & \multicolumn{1}{c|}{\textcolor{black}{0.270}} & \textcolor{black}{0.164} & \multicolumn{1}{c|}{\textcolor{black}{0.261}} & 0.167 & 0.263 & 0.203 & 0.298  & 0.166 & 0.263 & \textbf{0.161} & \textbf{0.252} \\

\hline

\multicolumn{1}{l|}{\multirow{5}{*}{\textbf{Traffic}}} 
& 96  & \textcolor{black}{0.379} & \textcolor{black}{0.274} & \textcolor{black}{0.383} & \multicolumn{1}{c|}{\textcolor{black}{0.280}} & \textcolor{black}{0.380} & \multicolumn{1}{c|}{\textcolor{black}{0.277}} & 0.388 & 0.282 & \underline{0.367} & \underline{0.288} & 0.410 & 0.282 & \textbf{0.360} & \textbf{0.249}   \\
\multicolumn{1}{l|}{}                       
& 192 & \textcolor{black}{0.397} & \underline{\textcolor{black}{0.282}} & \textcolor{black}{0.399} & \multicolumn{1}{c|}{\textcolor{black}{0.294}} & \textcolor{black}{0.399} & \multicolumn{1}{c|}{\textcolor{black}{0.288}} & 0.407 & 0.290 & \textbf{0.378} & 0.293 & 0.423 & 0.287 & \underline{0.379} & \textbf{0.256}    \\
\multicolumn{1}{l|}{}                       
& 336 & \textcolor{black}{0.407} & \underline{\textcolor{black}{0.289}} & \textcolor{black}{0.411} & \multicolumn{1}{c|}{\textcolor{black}{0.306}} & \textcolor{black}{0.408} & \multicolumn{1}{c|}{\textcolor{black}{0.290}} & 0.412 & 0.294 & \textbf{0.389} & 0.294 & 0.436 & 0.296 & \underline{0.392} & \textbf{0.264}    \\ 
\multicolumn{1}{l|}{}                      
& 720& \textcolor{black}{0.440} & \underline{\textcolor{black}{0.301}} & \textcolor{black}{0.448} & \multicolumn{1}{c|}{\textcolor{black}{0.319}} & \textcolor{black}{0.445} & \multicolumn{1}{c|}{\textcolor{black}{0.308}} & 0.450 & 0.312 & \textbf{0.401} & 0.304 & 0.466 & 0.315 & \underline{0.432} & \textbf{0.286}   \\
\multicolumn{1}{l|}{}                       
& Avg. & \textcolor{black}{0.405} & \underline{\textcolor{black}{0.286}} & \textcolor{black}{0.440} & \multicolumn{1}{c|}{\textcolor{black}{0.301}} & \textcolor{black}{0.408} & \multicolumn{1}{c|}{\textcolor{black}{0.290}} & 0.414 & 0.294 & \textbf{0.389} & 0.295  & 0.433 & 0.295 & \underline{0.390} & \textbf{0.263}  \\

\hline

\multicolumn{1}{l|}{\multirow{5}{*}{\textbf{ETTh1}}} 

& 96 & \textbf{\textcolor{black}{0.366}} & \textbf{\textcolor{black}{0.396}} & \textcolor{black}{0.380} & \multicolumn{1}{c|}{\textcolor{black}{0.406}} & \textcolor{black}{0.383} & \multicolumn{1}{c|}{\textcolor{black}{0.410}} & 0.379 & \underline{0.402} & 0.395 & 0.420 & \underline{0.367} & \textbf{0.396} & 0.379 & 0.407  \\
\multicolumn{1}{l|}{}                       
& 192 & \textbf{\textcolor{black}{0.401}} & \underline{\textcolor{black}{0.420}} & \textcolor{black}{0.426} & \multicolumn{1}{c|}{\textcolor{black}{0.438}} & \textcolor{black}{0.419} & \multicolumn{1}{c|}{\textcolor{black}{0.435}} & \underline{0.415} & 0.424 & 0.427 & 0.441 & \textbf{0.401} & \textbf{0.419} & 0.428 & 0.442  \\ 
\multicolumn{1}{l|}{}                       
& 336 & \textbf{\textcolor{black}{0.412}} & \textbf{\textcolor{black}{0.431}} & \textcolor{black}{0.437} & \textcolor{black}{0.451} & \underline{\textcolor{black}{0.426}} & \underline{{\textcolor{black}{0.440}}} & 0.435 & \underline{0.440} & 0.445 & 0.457 & 0.434 & 0.449 & 0.465 & 0.465   \\
\multicolumn{1}{l|}{}                      
& 720 & \underline{\textcolor{black}{0.440}} & \underline{\textcolor{black}{0.458}} & \textcolor{black}{0.515} & \multicolumn{1}{c|}{\textcolor{black}{0.509}} & \textbf{\textcolor{black}{0.428}} & \textbf{\textcolor{black}{0.456}} & 0.441 & 0.459 & 0.537  & 0.530 & 0.472 & 0.493 & 0.504 & 0.500  \\
\multicolumn{1}{l|}{}                       
& Avg. & \textbf{\textcolor{black}{0.406}} & \textbf{\textcolor{black}{0.427}} & \textcolor{black}{0.439} & \multicolumn{1}{c|}{\textcolor{black}{0.451}} & \underline{\textcolor{black}{0.414}} & \multicolumn{1}{c|}{\textcolor{black}{0.435}} & 0.418 & \underline{0.431} & 0.451 & 0.462 &  0.418 & 0.439 & 0.444 & 0.453   \\ 

\hline

\multicolumn{1}{l|}{\multirow{5}{*}{\textbf{ETTh2}}} 

& 96 & \textbf{\textcolor{black}{0.278}} & \textbf{\textcolor{black}{0.340}}  &  \textcolor{black}{0.306} & \multicolumn{1}{c|}{\textcolor{black}{0.362}} & \textcolor{black}{0.297} & \multicolumn{1}{c|}{\textcolor{black}{0.357}} & \underline{0.289} & \underline{0.347} & 0.304 & 0.360 & 0.301 & 0.367 & 0.296 & 0.353  \\
\multicolumn{1}{l|}{}                       
& 192 & \textbf{\textcolor{black}{0.346}} & \textbf{\textcolor{black}{0.385}} & \textbf{\textcolor{black}{0.346}} & \textbf{\textcolor{black}{0.385}} & \underline{\textcolor{black}{0.349}} & \underline{\textcolor{black}{0.390}} & 0.358 & 0.392 & 0.377 & 0.403 & 0.394 & 0.427 & 0.382 & 0.404  \\
\multicolumn{1}{l|}{}                       
& 336 & \textbf{\textcolor{black}{0.367}} & \textbf{\textcolor{black}{0.406}} & \textcolor{black}{0.393} & \multicolumn{1}{c|}{\textcolor{black}{0.422}} & \underline{\textcolor{black}{0.373}} & \underline{\textcolor{black}{0.408}} & 0.383 & 0.414 & 0.405 & 0.429 & 0.506 & 0.495 & 0.402 & 0.425  \\
\multicolumn{1}{l|}{}                      
& 720 & \underline{\textcolor{black}{0.400}} & \underline{\textcolor{black}{0.436}} & \textbf{\textcolor{black}{0.397}} & \textbf{\textcolor{black}{0.433}} & \underline{\textcolor{black}{0.400}} & \underline{\textcolor{black}{0.436}} & 0.438 & 0.456 & 0.443 & 0.464 & 0.805 & 0.635 & 0.444 & 0.465  \\
\multicolumn{1}{l|}{}                       
& Avg. & \textbf{\textcolor{black}{0.347}} & \textbf{\textcolor{black}{0.391}} & \textcolor{black}{0.360} & \multicolumn{1}{c|}{\textcolor{black}{0.400}} & \underline{\textcolor{black}{0.355}} & \underline{\textcolor{black}{0.398}} & 0.367 & 0.402 & 0.382 & 0.414 & 0.502 & 0.481 & 0.381 & 0.411  \\ 
\hline
\multicolumn{1}{l|}{\multirow{5}{*}{\textbf{ETTm1}}} 

& 96  & \textbf{\textcolor{black}{0.288}} & \textbf{\textcolor{black}{0.346}} &  \textcolor{black}{0.311} & \multicolumn{1}{c|}{\textcolor{black}{0.365}} & \underline{\textcolor{black}{0.291}} & \textbf{\textcolor{black}{0.346}} & 0.296 & 0.353 & 0.312 & 0.366 & 0.304 & \underline{0.348} & 0.303 & 0.351  \\
\multicolumn{1}{l|}{}                       
& 192 & \textbf{\textcolor{black}{0.323}} & \textbf{\textcolor{black}{0.365}} & \textcolor{black}{0.364} & \multicolumn{1}{c|}{\textcolor{black}{0.395}} & \textcolor{black}{0.336} & \multicolumn{1}{c|}{\textcolor{black}{0.373}} & \underline{0.335} & 0.373 & 0.347 & 0.385 & 0.336 & \underline{0.367} & 0.341 & 0.376 \\
\multicolumn{1}{l|}{}                       
& 336 & \textbf{\textcolor{black}{0.359}} & \underline{\textcolor{black}{0.390}}  & \textcolor{black}{0.369} & \multicolumn{1}{c|}{\textcolor{black}{0.398}} & \underline{\textcolor{black}{0.362}} & \underline{\textcolor{black}{0.390}} & 0.369 & 0.394 & 0.379 & 0.404 & 0.368 & \textbf{0.387} & 0.377 & 0.401   \\
\multicolumn{1}{l|}{}                      
& 720 & \textbf{\textcolor{black}{0.403}} & \textbf{\textcolor{black}{0.418}}  & \textcolor{black}{0.416} & \multicolumn{1}{c|}{\textcolor{black}{0.425}} & \underline{\textcolor{black}{0.410}} & \underline{\textcolor{black}{0.421}} & 0.418 & 0.424 & 0.441 & 0.442 & 0.421 & \textbf{0.418} & 0.431 & 0.436   \\
\multicolumn{1}{l|}{}                       
& Avg. & \textbf{\textcolor{black}{0.343}} & \textbf{\textcolor{black}{0.379}} & \textcolor{black}{0.365} & \multicolumn{1}{c|}{\textcolor{black}{0.395}} & \underline{\textcolor{black}{0.349}} & \underline{\textcolor{black}{0.382}} & 0.355 & 0.386 & 0.370 & 0.399 & 0.357 & 0.389 & 0.363 & 0.391  \\ 
\hline
\multicolumn{1}{l|}{\multirow{5}{*}{\textbf{ETTm2}}} 

& 96  & \textbf{\textcolor{black}{0.165}} & \textbf{\textcolor{black}{0.257}} & \textcolor{black}{0.170} & \underline{\textcolor{black}{0.262}} & \textcolor{black}{0.184} & \multicolumn{1}{c|}{\textcolor{black}{0.275}} & 0.170 & 0.264 & 0.179 & 0.271 & \underline{0.168} & 0.263 & 0.173 & 0.262   \\
\multicolumn{1}{l|}{}                       
& 192 & \textbf{\textcolor{black}{0.222}} & \textbf{\textcolor{black}{0.299}} & \underline{\textcolor{black}{0.229}} & \multicolumn{1}{c|}{\textcolor{black}{0.303}} & \textcolor{black}{0.238} & \multicolumn{1}{c|}{\textcolor{black}{0.310}} & 0.231 & 0.306 & 0.242 & 0.313 & \underline{0.229} & 0.310 & 0.231 & \underline{0.300}   \\
\multicolumn{1}{l|}{}                       
& 336 & \textbf{\textcolor{black}{0.277}} & \textbf{\textcolor{black}{0.330}} & \textcolor{black}{0.281} & \underline{\textcolor{black}{0.335}} &\textcolor{black}{0.286} & \multicolumn{1}{c|}{\textcolor{black}{0.340}} & \underline{0.280} & 0.339 & 0.288 & 0.344 & 0.289 & 0.352 & 0.292 & 0.345   \\ 
\multicolumn{1}{l|}{}                      
& 720 &  \textbf{\textcolor{black}{0.363}} & \textbf{\textcolor{black}{0.390}} & \textcolor{black}{0.379} & \multicolumn{1}{c|}{\textcolor{black}{0.403}} & \textcolor{black}{0.379} & \multicolumn{1}{c|}{\textcolor{black}{0.403}} & 0.373 & 0.402 & 0.378 & 0.397 & 0.416 & 0.437 & \underline{0.371} & \underline{0.394}   \\
\multicolumn{1}{l|}{}                       
& Avg. & \textbf{\textcolor{black}{0.257}} & \textbf{\textcolor{black}{0.319}} & \underline{\textcolor{black}{0.264}} & \underline{\textcolor{black}{0.325}} &  \textcolor{black}{0.271} & \multicolumn{1}{c|}{\textcolor{black}{0.332}} & 0.265 & 0.328 & 0.272 & 0.331 & 0.275 & 0.340 & 0.267 & \underline{0.325} \\

\bottomrule

\end{tabular}
\label{tab:long-term-forecast}
\end{table*}

\subsection{Long-term Forecasting}

\textbf{Setup.} For long-term forecasting, we evaluate the effectiveness of $S^2$IP-LLM on Weather, Electricity, Traffic, and four ETT datasets (\textit{i.e.}, ETTh1, ETTh2, ETTm1, and ETTm2), which have been widely adopted as benchmarking datasets for long-term forecasting tasks. Details of these datasets are shown in Appendix~\ref{A3}, Table \ref{data_stats}. The input time series length is 512, and we evaluate the performance on four different horizons $\{ 96, 192, 336, 720\}$. The evaluation metrics include the mean square error (MSE) and the mean absolute error (MAE).

\textbf{Results.} We compare the forecasting results of $S^2$IP-LLM to 6 selected baselines in Table \ref{tab:long-term-forecast}. Due to the space limitation, the comparisons with the other 6 baselines are provided in Appendix~\ref{sec:appendix:long-term} and Table~\ref{tab:long-term-forecast-transformer}. We can observe that LLMs based forecasting methods, \textit{i.e.}, Time-LLM and OFA, generally achieve better performance than other baseline methods. This should be attributed to the prevalent expressibility of LLMs and their associated prompt-tuning and fine-tuning strategies, respectively. Moreover, most of the time, $S^2$IP-LLM outperforms Time-LLM and OFA over 7 different datasets. This is because (1) the unique way $S^2$IP-LLM tokenized the input time series data can yield better time series representations, and (2) the semantic space informed prompting can help further enhance the time series representation which will be further demonstrated in Section 4.5.


\subsection{Short-term Forecasting}

\textbf{Setup.} We also evaluate the effectiveness of $S^2$IP-LLM with the short-term forecasting setting based on the M4 datasets~\citep{makridakis2018m4}. It contains a collection of marketing data that are sampled at different frequencies. Details of the datasets can be found in Appendix~\ref{A3}. The prediction horizons are significantly shorter than the long-term forecasting setting and are set between 6 and 48. The input lengths are twice the prediction horizons, similar to the experiment setting in \citep{jin2023time,zhou2023onefitsall}. The evaluation metrics for short-term forecasting are symmetric mean absolute percentage error (SMAPE), mean absolute scaled error (MASE), and overall weighted average (OWA). The details of these evaluation metrics are provided in Appendix~\ref{A4}.

\textbf{Results.} Table~\ref{tab:short-term-forecast-summary} summarizes the short-term forecasting results and the full experiment results are shown in Appendix Appendix~\ref{appendix:short-term}, Table ~\ref{tab:short-term-forecast-full}. We observe that $S^2$IP-LLM outperforms all other baselines by a large margin and is slightly better than PatchTST. This could attribute to the tokenization design as well as the semantic space informed prompting within $S^2$IP-LLM.


\begin{table*}[ht]
\footnotesize
\centering
\caption{Short-term time series forecasting results on M4 datasets. The forecasting horizons are in [6, 48] and the
three rows provided are weighted averaged from all datasets under different sampling intervals. A lower value
indicates better performance. 
Detailed short-term forecasting results are in Appendix~\ref{appendix:short-term}, Table~\ref{tab:short-term-forecast-full}
%
%
} 
\setlength{\tabcolsep}{3.0pt}
\begin{tabular}{cc|ccccccccccc}
\hline\hline
\multicolumn{2}{c|}{Methods} & \textbf{$\mathbf{S^2}$IP-LLM} & \textbf{Time-LLM(G)} & \textbf{OFA} & \textbf{iTransformer} & \textbf{Dlinear} & \textbf{PatchTST} & \textbf{TimesNet} & \textbf{FEDformer} & \textbf{Autoformer}  \\ 
\hline

\multirow{3}{*}{Avg.} & \textbf{SMAPE} & \textbf{\textcolor{black}{12.021}} & \textcolor{black}{12.494} & 12.690 & 12.142 & 13.639 & \underline{12.059} & 12.880 & 13.160 & 12.909   \\ 

& \textbf{MASE} & \textbf{\textcolor{black}{1.612}} & \textcolor{black}{1.731} & 1.808 & 1.631 & 2.095 & \underline{1.623} & 1.836 & 1.775 & 1.771 \\

& \textbf{OWA} & \textbf{\textcolor{black}{0.857}} & \textcolor{black}{0.913} & 0.94 & 0.874 & 1.051 & \underline{0.869}  & 0.955 & 0.949 & 0.939   \\ 
\hline
\end{tabular}
\label{tab:short-term-forecast-summary}
\end{table*}\vspace{5mm}

\begin{table*}[htbp]
\caption{{Long-term forecasting results for \{96, 192, 336, 720\} horizons. A lower value indicates a better performance.
Few-shot learning on 10\% training data setting. All results are averaged from four forecasting horizons\{96, 192, 336, 720\}. Detailed results are in Appendix C,Table~\ref{tab:long-term-forecast_few10}.}}
\footnotesize
\centering
\setlength{\tabcolsep}{4.8pt}
\begin{tabular}{cc|cc|cc|cc|cc|cc|cc|cc}
\hline\hline
\multicolumn{2}{c|}{Methods} &
  \multicolumn{2}{c|}{\textbf{$\mathbf{S^2}$IP-LLM}} &
  \multicolumn{2}{c|}{\textbf{Time-LLM(G)}}&
  \multicolumn{2}{c|}{\textbf{OFA}} &
  \multicolumn{2}{c|}{\textbf{iTransformer}} &
  \multicolumn{2}{c|}{\textbf{Dlinear}} &
  \multicolumn{2}{c|}{\textbf{PatchTST}} &
  \multicolumn{2}{c}{\textbf{TimesNet}}    \\ \hline
\multicolumn{2}{c|}{Metric} &
  \multicolumn{1}{c}{MSE} &
  \multicolumn{1}{c|}{MAE} &
  \multicolumn{1}{c}{MSE} &
  \multicolumn{1}{c|}{MAE} &
  \multicolumn{1}{c}{MSE} &
  \multicolumn{1}{c|}{MAE} &
  \multicolumn{1}{c}{MSE} &
  \multicolumn{1}{c|}{MAE} &
  \multicolumn{1}{c}{MSE} &
  \multicolumn{1}{c|}{MAE} &
  \multicolumn{1}{c}{MSE} &
  \multicolumn{1}{c|}{MAE} &
  \multicolumn{1}{c}{MSE} &
  \multicolumn{1}{c}{MAE} \\ \hline

\multicolumn{2}{l|}{\textbf{Weather}} & \textbf{\textcolor{black}{0.233}} & \textbf{\textcolor{black}{0.272}} & \underline{\textcolor{black}{0.237}} & \underline{\textcolor{black}{0.275}} & 0.238 & 0.275 & 0.308 & 0.338 & 0.241 & 0.283 & 0.242 & 0.279 & 0.279 & 0.301 \\
\hline

\multicolumn{2}{l|}{\textbf{Electricity}} & \textbf{\textcolor{black}{0.175}} & \underline{\textcolor{black}{0.271}} & \textcolor{black}{0.177} & \textcolor{black}{0.273} & \underline{0.176} & \textbf{0.269} & 0.196  &  0.293 &    0.180 & 0.280 & 0.180 & 0.273 & 0.323 & 0.392  \\
\hline

\multicolumn{2}{l|}{\textbf{Traffic}} & \textbf{\textcolor{black}{0.427}} & \underline{\textcolor{black}{0.307}} & \textcolor{black}{0.429} & \underline{\textcolor{black}{0.307}} & 0.440 & 0.310 &  0.495 &  0.361 & 0.447 & 0.313 & 0.430 & \textbf{0.305} & 0.951 & 0.535 \\
\hline
\multicolumn{2}{l|}{\textbf{ETTh1}} & \underline{\textcolor{black}{0.593}} & \underline{\textcolor{black}{0.529}} & \textcolor{black}{0.785} & \textcolor{black}{0.553} & \textbf{0.590} & \textbf{0.525} & 0.910 & 0.860  & 0.691 & 0.600 & 0.633 & 0.542 & 0.869 & 0.628  \\
\hline

\multicolumn{2}{l|}{\textbf{ETTh2}}& \textcolor{black}{0.419} & \textcolor{black}{0.439} & \textcolor{black}{0.424} & \textcolor{black}{0.441} & \textbf{0.397} & \textbf{0.421} & 0.489 & 0.483  & 0.605 & 0.538 & \underline{0.415} & \underline{0.431} & 0.479 & 0.465  \\
\hline

\multicolumn{2}{l|}{\textbf{ETTm1}}& \underline{\textcolor{black}{0.455}} & \underline{\textcolor{black}{0.435}} & \textcolor{black}{0.487} & \textcolor{black}{0.461} & 0.464 & 0.441 & 0.728 & 0.565 & \textbf{0.411} & \textbf{0.429} & 0.501 & 0.466 & 0.677 & 0.537  \\
\hline

\multicolumn{2}{l|}{\textbf{ETTm2}} & \textbf{\textcolor{black}{0.284}} & \textbf{\textcolor{black}{0.332}} & \textcolor{black}{0.305} & \textcolor{black}{0.344} & \underline{0.293} & \underline{0.335} & 0.336 & 0.373 & 0.316 & 0.368 & 0.296 & 0.343 & 0.320 & 0.353 \\

\bottomrule

\end{tabular}
\label{tab:few10}
\end{table*}\vspace{5mm}

\begin{table*}[htbp]
\caption{Long-term forecasting results for \{96, 192, 336, 720\} horizons. A lower value indicates a better performance.
Few-shot learning on 5\% training data setting. All results are averaged from four forecasting horizons\{96, 192, 336, 720\}. Detailed results are in Appendix C, Table~\ref{tab:long-term-forecast_few5}. }

\footnotesize
\centering
\setlength{\tabcolsep}{4.8pt}
\begin{tabular}{cc|cc|cc|cc|cc|cc|cc|cc}
\hline\hline
\multicolumn{2}{c|}{Methods} &
  \multicolumn{2}{c|}{\textbf{$\mathbf{S^2}$IP-LLM}} &
  \multicolumn{2}{c|}{\textbf{Time-LLM(G)}}&
  \multicolumn{2}{c|}{\textbf{OFA}} &
  \multicolumn{2}{c|}{\textbf{iTransformer}} &
  \multicolumn{2}{c|}{\textbf{Dlinear}} &
  \multicolumn{2}{c|}{\textbf{PatchTST}} &
  \multicolumn{2}{c}{\textbf{TimesNet}}  \\ \hline
\multicolumn{2}{c|}{Metric} &
  \multicolumn{1}{c}{MSE} &
  \multicolumn{1}{c|}{MAE} &
  \multicolumn{1}{c}{MSE} &
  \multicolumn{1}{c|}{MAE} &
  \multicolumn{1}{c}{MSE} &
  \multicolumn{1}{c|}{MAE} &
  \multicolumn{1}{c}{MSE} &
  \multicolumn{1}{c|}{MAE} &
  \multicolumn{1}{c}{MSE} &
  \multicolumn{1}{c|}{MAE} &
  \multicolumn{1}{c}{MSE} &
  \multicolumn{1}{c|}{MAE} &
  \multicolumn{1}{c}{MSE} &
  \multicolumn{1}{c}{MAE}  \\ \hline

\multicolumn{2}{l|}{\textbf{Weather}} & \textbf{\textcolor{black}{0.260}} & \textbf{\textcolor{black}{0.297}} & \textcolor{black}{0.264} & \textcolor{black}{0.301} & \underline{0.263} & \underline{0.301} & 0.309 &  0.339 & \underline{0.263} & 0.308 & 0.269 & 0.303 & 0.298 & 0.318  \\
\hline

\multicolumn{2}{l|}{\textbf{Electricity}} & \textcolor{black}{0.179} & \underline{\textcolor{black}{0.275}} & \textcolor{black}{0.181} & \textcolor{black}{0.279} & \underline{0.178} & \textbf{0.273} & 0.201 & 0.296 & \textbf{0.176} & \underline{0.275} & 0.181 & 0.277 & 0.402 & 0.453 
\\
\hline

\multicolumn{2}{l|}{\textbf{Traffic}} & \underline{\textcolor{black}{0.420}} & \underline{\textcolor{black}{0.299}} & \textcolor{black}{0.423} & \textcolor{black}{0.302}& 0.434 & 0.305 & 0.450 & 0.324 & 0.450 & 0.317 & \textbf{0.418} & \textbf{0.296} & 0.867 & 0.493 
\\
\hline

\multicolumn{2}{l|}{\textbf{ETTh1}} &  \textbf{\textcolor{black}{0.650}} & \textbf{\textcolor{black}{0.550}} & \textcolor{black}{0.891} & \textcolor{black}{0.627} & \underline{0.681} & \underline{0.560} & 1.070 & 0.710 & 0.750 & 0.611 & 0.694 & 0.569 & 0.925 & 0.647 \\ 
\hline

\multicolumn{2}{l|}{\textbf{ETTh2}} & \textbf{\textcolor{black}{0.380}} & \textbf{\textcolor{black}{0.413}} & \textcolor{black}{0.581} & \textcolor{black}{0.519} & \underline{0.400} & \underline{0.433} & 0.488 & 0.475 & 0.694 & 0.577 & 0.827 & 0.615 & 0.439 & 0.448 \\
\hline

\multicolumn{2}{l|}{\textbf{ETTm1}} & \underline{\textcolor{black}{0.455}} & \underline{\textcolor{black}{0.446}} & \textcolor{black}{0.524} & \textcolor{black}{0.479} & 0.472 & 0.450 & 0.784 & 0.596 & \textbf{0.400} & \textbf{0.417} & 0.526 & 0.476 & 0.717 & 0.561 \\
\hline

\multicolumn{2}{l|}{\textbf{ETTm2}} & \textbf{\textcolor{black}{0.296}} & \textbf{\textcolor{black}{0.342}} & \textcolor{black}{0.325} & \textcolor{black}{0.361} & \underline{0.308} & \underline{0.346} & 0.356 & 0.388 & 0.399 & 0.426 & 0.314 & 0.352 & 0.344 & 0.372  
\\

\bottomrule

\end{tabular}
\label{tab:few5}
\end{table*}

\subsection{Few-shot Forecasting}

\textbf{Setup.} We follow the experimental settings in~\citet{zhou2023onefitsall} to evaluate the performance in the few-shot forecasting setting, which allows us to examine whether the model can generate accurate forecasting with limited training data. We use the first 5\% and 10\% of the training data in these experiments.

\textbf{Results.} To ensure a fair comparison in the long-term forecasting setting, we summarize the few-shot learning experiment results under 10\% and 5\% training data in Table~\ref{tab:few10} and Table~\ref{tab:few5}, respectively. We also report the full experiment results in Table~\ref{tab:long-term-forecast_few10} and Table~\ref{tab:long-term-forecast_few5}  of Appendix~\ref{appendix:few-shot}, respectively.
When trained with only 10\% of the data, $S^2$IP-LLM typically ranks as either the best or the second-best compared to other baseline models across different datasets. Meanwhile, we also observe that LLMs based methods,  $S^2$IP-LLM, Time-LLM, and OFA significantly outperform other baseline methods. This is because other baseline methods are trained from scratch and they only have limited training data in this case. On the other hand, LLMs based methods can adapt/align the pre-trained knowledge with the time series embedding to enhance its representation. Even with only 5\% of training data, $S^2$IP-LLM still exhibits, if not superior, comparable performance to time-LLM and OFA.

\subsection{Ablation Studies and Parameter Sensitivity}

\begin{figure*}[!h]
\begin{center}
\centerline{\includegraphics[width=\textwidth]{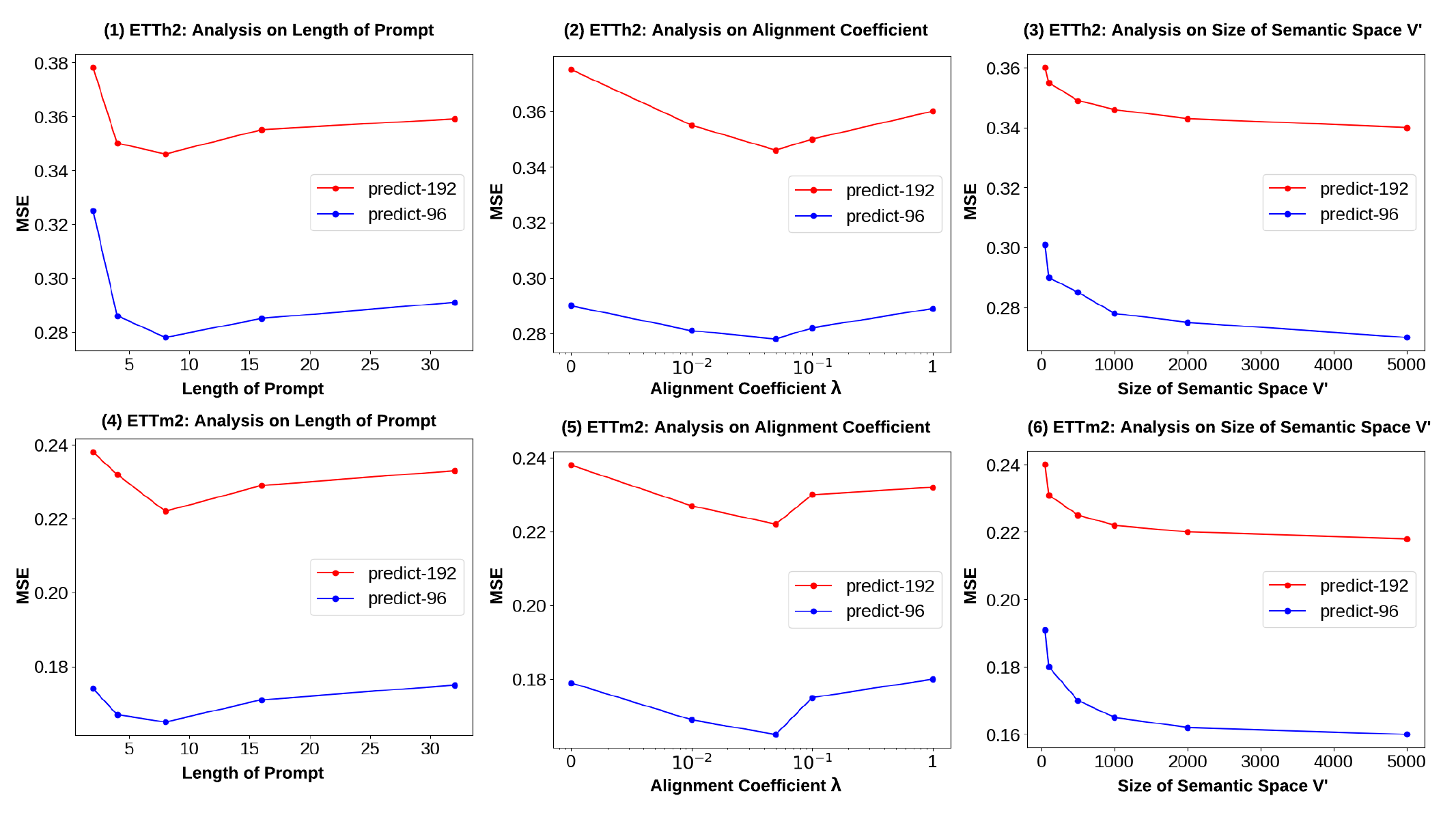}}
\caption{Parameter sensitivity analysis in predicting 96 and 192 steps: \textcolor{black}{ (1) and (4) show the effect of prompt length on ETTh2 and ETTm2 datasets; (2) and (5) show the effect of alignment coefficient $\lambda$ on ETTh2 and ETTm2 datasets; (3) and (6) show the effect of semantic space size $V^{\prime}$ on ETTh2 and ETTm2 datasets.}}

\label{parameter}
\end{center}
\end{figure*}

\textcolor{black}{We conduct ablation studies on the ETTh2 and ETTm2 datasets to evaluate the parameter sensitivity for $S^2$IP-LLM.} \textcolor{black}{Figure \ref{parameter} (1) and (4)} presents the experiment results when the length of the prompt varies on ETTh2 and ETTm2, respectively. Within a limited range, \textit{i.e.} 2 to 8, an increase in the prompt length tends to improve the forecasting performance. However, excessive prompt length, such as lengths of 16 or 32, results in a significant decline in the forecasting accuracy. A similar pattern can be observed in the hyperparameter analysis of the $\lambda$, which controls the strength of alignment. As shown in Figure \ref{parameter} (2) and (5), when $\lambda$ varies from 0 to 0.05, slightly larger $\lambda$ is beneficial for representation learning within the joint space, showing better forecasting results. On the other hand, larger $\lambda$ tends to lead to indistinguishable time series representation and the forecasting performance will thus decrease. Finally, Figure \ref{parameter} (3) and (6) show the effects of choosing different numbers of semantic anchors. Generally, an increased number of semantic anchors improves the forecasting results. We conjecture that the small number hinders the learning of highly representative semantic anchors in the joint space and thus will generate less informed prompts for time series embedding. We visualize the prompted time series embeddings with different numbers of semantic anchors in Appendix~\ref{Appendix: ablation}, Figure \ref{tsne-vis_3}. We notice that a smaller quantity of semantic anchors leads to a less dispersed distribution in the joint space, indicating that the generated prompts could be less informative for time series embedding. \textcolor{black}{We also perform ablation studies by incrementally adding the ``alignment \& prompting" and ``decomposition" modules. In Appendix~\ref{Appendix: ablation} Table~\ref{Ablation}, we observe the forecasting performance increases when we sequentially activate the prompting \& alignment component and the decomposition component, which implies the importance of these modules in $S^2$IP-LLM.}

\subsection{Qualitative Analysis}

\begin{figure}[t]
\begin{center}
\centerline{\includegraphics[width=\columnwidth]{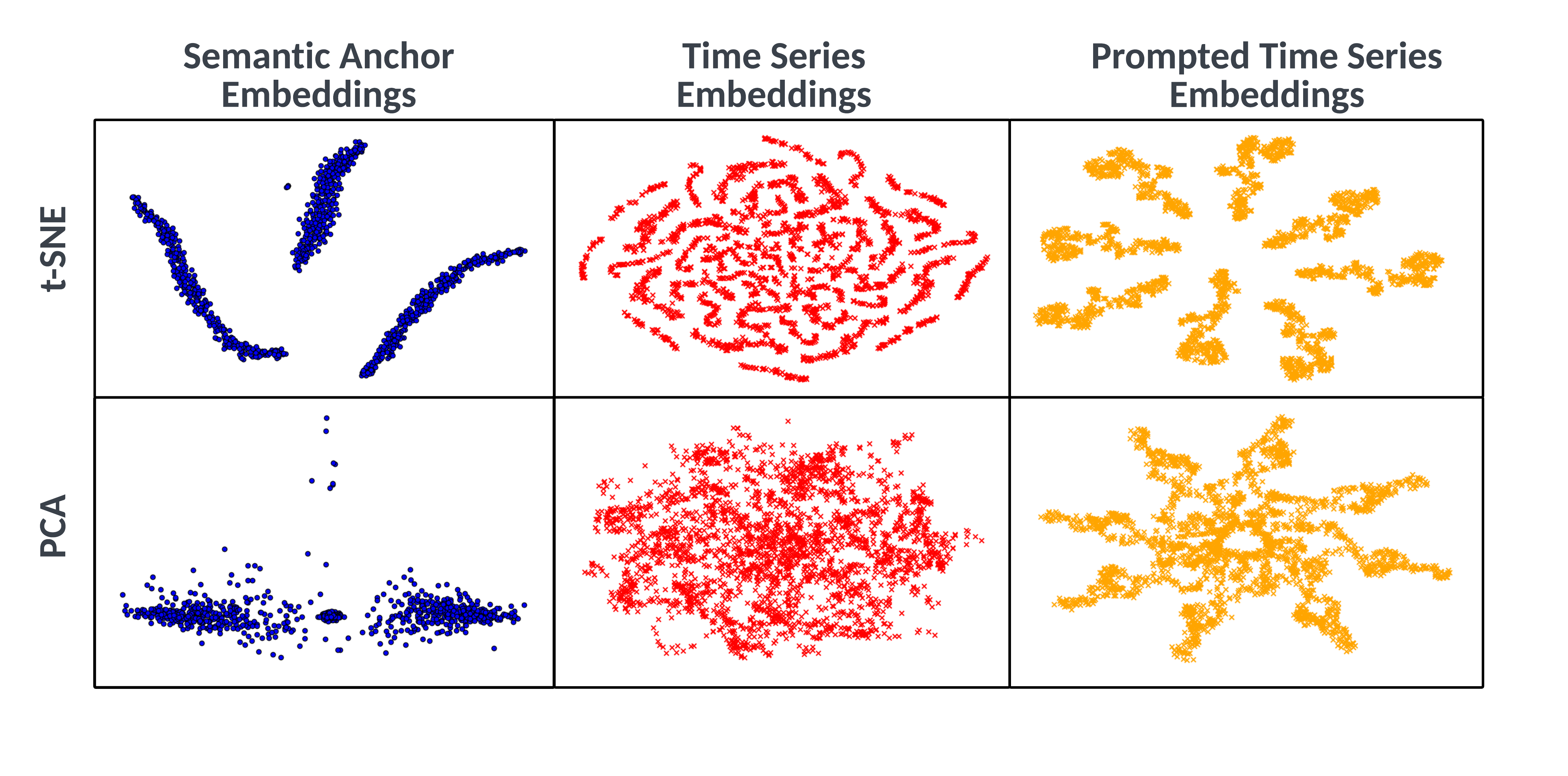}}

\caption{The t-SNE and PCA plots of embeddings space: \textbf{\textcolor{blue}{blue}}: semantic anchor embeddings; \textbf{\textcolor{red}{red}}: time series embeddings;
\textbf{\textcolor{orange}{orange}}: prefix-prompted time series embeddings}
\label{tsne-vis}
\end{center}
\end{figure}

In this section, we perform a qualitative analysis of how semantic space informed prompting can facilitate time series representation. Figure~\ref{tsne-vis} shows the visualization of learned semantic anchor embeddings, time series embeddings, and the prompted time series embeddings. The semantic anchor embeddings from the pre-trained language model show distinct clusters, suggesting a robust and differentiated embedding space. In contrast, the raw time series embeddings reveal a more spread-out and less clustered pattern, suggesting that before the alignment, the time series representation is comparatively less informative. After the alignment, the prompted time series embeddings show a clear clustered pattern, suggesting that by aligning with the semantic anchors, time series representation becomes more distinguishable in the joint space.

We also provide the visualizations of prompted time series embeddings under different hyperparameters (when $\lambda$ varies). Within a smaller range, the increase of $\lambda$ appears to enhance the separation of time series embeddings, indicating a more distinct and informative representation. However, as $\lambda$ becomes excessively large, we observe a significant decline in the clustering quality of the prompted time series embeddings, which suggests that beyond a threshold, a higher $\lambda$ value leads to less informative embeddings.

\begin{figure}[t]
\begin{center}
\centerline{\includegraphics[width=\columnwidth]{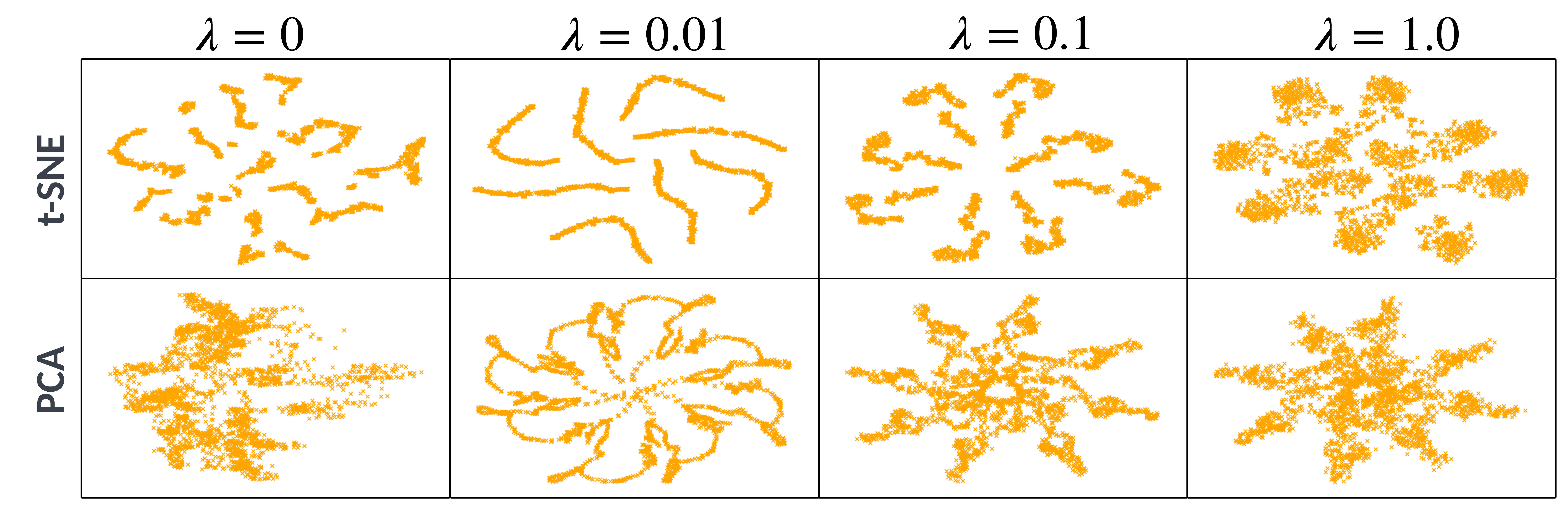}}
\caption{The t-SNE and PCA plots prefix-prompted time series embeddings with different $\lambda$} 
\label{tsne-vis_2}
\end{center}
\end{figure}

\section{Conclusion}

In this paper, we present $S^2$IP-LLM, a novel framework for time series forecasting utilizing pre-trained language models. $S^2$IP-LLM introduces a time series tokenization module that provides expressive local contexts by the concatenation of decomposed time series patches. It creates informative joint space by aligning time series contexts with semantics anchors derived from the pre-trained word token embeddings. The selected aligned semantic anchors are retrieved as prompt indicators to enhance the time series representation and facilitate underlying forecasting tasks. Our thorough empirical studies justified the effectiveness of $S^2$IP-LLM.

\section*{Impact Statement}

This work introduces significant advancements in time series forecasting, leveraging the power of pre-trained language models and semantic information. The broader impact of this work can be multifaceted. It may enhance decision-making in critical domains such as finance, healthcare, and environmental monitoring by providing more accurate and reliable forecasts and could lead to better resource allocation, improved patient care, and more effective responses to climate change. No ethical concerns must be considered. The social impacts are significant, as it has the potential to revolutionize our approach to complex time series data and the integration of emerging AI tools, including foundational models. It could change how we analyze and leverage time series data in various fields.


\begin{thebibliography}{67}
\providecommand{\natexlab}[1]{#1}
\providecommand{\url}[1]{\texttt{#1}}
\expandafter\ifx\csname urlstyle\endcsname\relax
  \providecommand{\doi}[1]{doi: #1}\else
  \providecommand{\doi}{doi: \begingroup \urlstyle{rm}\Url}\fi

\bibitem[Achiam \& et~al.(2023)Achiam and et~al.]{Achiam2023GPT4TR}
Achiam, O.~J. and et~al., S.~A.
\newblock Gpt-4 technical report.
\newblock 2023.
\newblock URL \url{https://api.semanticscholar.org/CorpusID:257532815}.

\bibitem[Anderson \& Kendall(1976)Anderson and Kendall]{Anderson1976TimeSeries2E}
Anderson, O.~D. and Kendall, M.~G.
\newblock Time-series. 2nd edn.
\newblock \emph{The Statistician}, 25:\penalty0 308, 1976.
\newblock URL \url{https://api.semanticscholar.org/CorpusID:134001785}.

\bibitem[Bao et~al.(2021)Bao, Dong, Piao, and Wei]{bao2021beit}
Bao, H., Dong, L., Piao, S., and Wei, F.
\newblock Beit: Bert pre-training of image transformers.
\newblock \emph{arXiv preprint arXiv:2106.08254}, 2021.

\bibitem[B{\"o}se et~al.(2017)B{\"o}se, Flunkert, Gasthaus, Januschowski, Lange, Salinas, Schelter, Seeger, and Wang]{bose2017probabilistic}
B{\"o}se, J.-H., Flunkert, V., Gasthaus, J., Januschowski, T., Lange, D., Salinas, D., Schelter, S., Seeger, M., and Wang, Y.
\newblock Probabilistic demand forecasting at scale.
\newblock \emph{Proceedings of the VLDB Endowment}, 10\penalty0 (12):\penalty0 1694--1705, 2017.

\bibitem[Brown et~al.(2020)Brown, Mann, Ryder, Subbiah, Kaplan, Dhariwal, Neelakantan, Shyam, Sastry, Askell, et~al.]{brown2020language}
Brown, T., Mann, B., Ryder, N., Subbiah, M., Kaplan, J.~D., Dhariwal, P., Neelakantan, A., Shyam, P., Sastry, G., Askell, A., et~al.
\newblock Language models are few-shot learners.
\newblock \emph{Advances in neural information processing systems}, 33:\penalty0 1877--1901, 2020.

\bibitem[Cao et~al.(2020)Cao, Wang, Duan, Zhang, Zhu, Huang, Tong, Xu, Bai, Tong, et~al.]{cao2020spectral}
Cao, D., Wang, Y., Duan, J., Zhang, C., Zhu, X., Huang, C., Tong, Y., Xu, B., Bai, J., Tong, J., et~al.
\newblock Spectral temporal graph neural network for multivariate time-series forecasting.
\newblock \emph{Advances in neural information processing systems}, 33:\penalty0 17766--17778, 2020.

\bibitem[Cao et~al.(2023)Cao, Jia, Arik, Pfister, Zheng, Ye, and Liu]{cao2023tempo}
Cao, D., Jia, F., Arik, S.~O., Pfister, T., Zheng, Y., Ye, W., and Liu, Y.
\newblock Tempo: Prompt-based generative pre-trained transformer for time series forecasting.
\newblock \emph{arXiv preprint arXiv:2310.04948}, 2023.

\bibitem[Challu et~al.(2023)Challu, Olivares, Oreshkin, Ramirez, Canseco, and Dubrawski]{challu2023nhits}
Challu, C., Olivares, K.~G., Oreshkin, B.~N., Ramirez, F.~G., Canseco, M.~M., and Dubrawski, A.
\newblock Nhits: Neural hierarchical interpolation for time series forecasting.
\newblock In \emph{Proceedings of the AAAI Conference on Artificial Intelligence}, volume~37, pp.\  6989--6997, 2023.

\bibitem[Chang et~al.(2023)Chang, Peng, and Chen]{chang2023llm4ts}
Chang, C., Peng, W.-C., and Chen, T.-F.
\newblock Llm4ts: Two-stage fine-tuning for time-series forecasting with pre-trained llms.
\newblock \emph{arXiv preprint arXiv:2308.08469}, 2023.

\bibitem[Cleveland et~al.(1990)Cleveland, Cleveland, McRae, and Terpenning]{cleveland1990stl}
Cleveland, R.~B., Cleveland, W.~S., McRae, J.~E., and Terpenning, I.
\newblock Stl: A seasonal-trend decomposition.
\newblock \emph{J. Off. Stat}, 6\penalty0 (1):\penalty0 3--73, 1990.

\bibitem[Courty \& Li(1999)Courty and Li]{courty1999timing}
Courty, P. and Li, H.
\newblock Timing of seasonal sales.
\newblock \emph{The Journal of Business}, 72\penalty0 (4):\penalty0 545--572, 1999.

\bibitem[Cui et~al.(2024)Cui, Ma, Cao, Ye, Zhou, Liang, Chen, Lu, Yang, Liao, et~al.]{cui2024survey}
Cui, C., Ma, Y., Cao, X., Ye, W., Zhou, Y., Liang, K., Chen, J., Lu, J., Yang, Z., Liao, K.-D., et~al.
\newblock A survey on multimodal large language models for autonomous driving.
\newblock In \emph{Proceedings of the IEEE/CVF Winter Conference on Applications of Computer Vision}, pp.\  958--979, 2024.

\bibitem[Deldari et~al.(2022)Deldari, Xue, Saeed, He, Smith, and Salim]{deldari2022beyond}
Deldari, S., Xue, H., Saeed, A., He, J., Smith, D.~V., and Salim, F.~D.
\newblock Beyond just vision: A review on self-supervised representation learning on multimodal and temporal data.
\newblock \emph{arXiv preprint arXiv:2206.02353}, 2022.

\bibitem[Devlin et~al.(2018)Devlin, Chang, Lee, and Toutanova]{devlin2018bert}
Devlin, J., Chang, M.-W., Lee, K., and Toutanova, K.
\newblock Bert: Pre-training of deep bidirectional transformers for language understanding.
\newblock \emph{arXiv preprint arXiv:1810.04805}, 2018.

\bibitem[Dimri et~al.(2020)Dimri, Ahmad, and Sharif]{dimri2020time}
Dimri, T., Ahmad, S., and Sharif, M.
\newblock Time series analysis of climate variables using seasonal arima approach.
\newblock \emph{Journal of Earth System Science}, 129:\penalty0 1--16, 2020.

\bibitem[Ethayarajh(2019)]{ethayarajh2019contextual}
Ethayarajh, K.
\newblock How contextual are contextualized word representations? comparing the geometry of bert, elmo, and gpt-2 embeddings.
\newblock \emph{arXiv preprint arXiv:1909.00512}, 2019.

\bibitem[Fawaz et~al.(2018)Fawaz, Forestier, Weber, Idoumghar, and Muller]{fawaz2018transfer}
Fawaz, H.~I., Forestier, G., Weber, J., Idoumghar, L., and Muller, P.-A.
\newblock Transfer learning for time series classification.
\newblock In \emph{2018 IEEE international conference on big data (Big Data)}, pp.\  1367--1376. IEEE, 2018.

\bibitem[Friedman(1962)]{friedman1962interpolation}
Friedman, M.
\newblock The interpolation of time series by related series.
\newblock \emph{Journal of the American Statistical Association}, 57\penalty0 (300):\penalty0 729--757, 1962.

\bibitem[Gao et~al.(2020)Gao, Song, Wen, Wang, Sun, and Xu]{gao2020robusttad}
Gao, J., Song, X., Wen, Q., Wang, P., Sun, L., and Xu, H.
\newblock Robusttad: Robust time series anomaly detection via decomposition and convolutional neural networks.
\newblock \emph{arXiv preprint arXiv:2002.09545}, 2020.

\bibitem[Garza \& Mergenthaler-Canseco(2023)Garza and Mergenthaler-Canseco]{garza2023timegpt}
Garza, A. and Mergenthaler-Canseco, M.
\newblock Timegpt-1.
\newblock \emph{arXiv preprint arXiv:2310.03589}, 2023.

\bibitem[Gu et~al.(2021)Gu, Goel, and R{\'e}]{gu2021efficiently}
Gu, A., Goel, K., and R{\'e}, C.
\newblock Efficiently modeling long sequences with structured state spaces.
\newblock \emph{arXiv preprint arXiv:2111.00396}, 2021.

\bibitem[He et~al.(2022)He, Chen, Xie, Li, Doll{\'a}r, and Girshick]{he2022masked}
He, K., Chen, X., Xie, S., Li, Y., Doll{\'a}r, P., and Girshick, R.
\newblock Masked autoencoders are scalable vision learners.
\newblock In \emph{Proceedings of the IEEE/CVF conference on computer vision and pattern recognition}, pp.\  16000--16009, 2022.

\bibitem[Houlsby et~al.(2019)Houlsby, Giurgiu, Jastrzebski, Morrone, De~Laroussilhe, Gesmundo, Attariyan, and Gelly]{houlsby2019parameter}
Houlsby, N., Giurgiu, A., Jastrzebski, S., Morrone, B., De~Laroussilhe, Q., Gesmundo, A., Attariyan, M., and Gelly, S.
\newblock Parameter-efficient transfer learning for nlp.
\newblock In \emph{International Conference on Machine Learning}, pp.\  2790--2799. PMLR, 2019.

\bibitem[Jiang et~al.(2024)Jiang, Pan, Zhang, Garg, Schneider, Nevmyvaka, and Song]{jiang2024empowering}
Jiang, Y., Pan, Z., Zhang, X., Garg, S., Schneider, A., Nevmyvaka, Y., and Song, D.
\newblock Empowering time series analysis with large language models: A survey.
\newblock \emph{arXiv preprint arXiv:2402.03182}, 2024.

\bibitem[Jin et~al.(2024)Jin, Wang, Ma, Chu, Zhang, Shi, Chen, Liang, Li, Pan, et~al.]{jin2023time}
Jin, M., Wang, S., Ma, L., Chu, Z., Zhang, J.~Y., Shi, X., Chen, P.-Y., Liang, Y., Li, Y.-F., Pan, S., et~al.
\newblock Time-llm: Time series forecasting by reprogramming large language models.
\newblock In \emph{International Conference on Learning Representations}, 2024.

\bibitem[Kim et~al.(2021)Kim, Kim, Tae, Park, Choi, and Choo]{kim2021reversible}
Kim, T., Kim, J., Tae, Y., Park, C., Choi, J.-H., and Choo, J.
\newblock Reversible instance normalization for accurate time-series forecasting against distribution shift.
\newblock In \emph{International Conference on Learning Representations}, 2021.

\bibitem[Lai et~al.(2018)Lai, Chang, Yang, and Liu]{lai2018modeling}
Lai, G., Chang, W.-C., Yang, Y., and Liu, H.
\newblock Modeling long-and short-term temporal patterns with deep neural networks.
\newblock In \emph{The 41st international ACM SIGIR conference on research \& development in information retrieval}, pp.\  95--104, 2018.

\bibitem[Lester et~al.(2021)Lester, Al-Rfou, and Constant]{lester2021power}
Lester, B., Al-Rfou, R., and Constant, N.
\newblock The power of scale for parameter-efficient prompt tuning.
\newblock \emph{arXiv preprint arXiv:2104.08691}, 2021.

\bibitem[Li et~al.(2022)Li, Arnold, Down, Barty, Blake, Chiang, Courtney, Waito, Trifunov, and Heddle]{li2022demand}
Li, N., Arnold, D.~M., Down, D.~G., Barty, R., Blake, J., Chiang, F., Courtney, T., Waito, M., Trifunov, R., and Heddle, N.~M.
\newblock From demand forecasting to inventory ordering decisions for red blood cells through integrating machine learning, statistical modeling, and inventory optimization.
\newblock \emph{Transfusion}, 62\penalty0 (1):\penalty0 87--99, 2022.

\bibitem[Li et~al.(2017)Li, Yu, Shahabi, and Liu]{li2017diffusion}
Li, Y., Yu, R., Shahabi, C., and Liu, Y.
\newblock Diffusion convolutional recurrent neural network: Data-driven traffic forecasting.
\newblock \emph{arXiv preprint arXiv:1707.01926}, 2017.

\bibitem[Li et~al.(2023)Li, Wang, Ding, and Chen]{li2023large}
Li, Y., Wang, S., Ding, H., and Chen, H.
\newblock Large language models in finance: A survey.
\newblock In \emph{Proceedings of the Fourth ACM International Conference on AI in Finance}, pp.\  374--382, 2023.

\bibitem[Liu et~al.(2023{\natexlab{a}})Liu, Ma, Yang, Zhou, Xia, Wang, Wen, and Sun]{liu2023sadi}
Liu, H., Ma, Z., Yang, L., Zhou, T., Xia, R., Wang, Y., Wen, Q., and Sun, L.
\newblock Sadi: A self-adaptive decomposed interpretable framework for electric load forecasting under extreme events.
\newblock In \emph{ICASSP 2023-2023 IEEE International Conference on Acoustics, Speech and Signal Processing (ICASSP)}, pp.\  1--5. IEEE, 2023{\natexlab{a}}.

\bibitem[Liu et~al.(2022)Liu, Wu, Wang, and Long]{liu2022non}
Liu, Y., Wu, H., Wang, J., and Long, M.
\newblock Non-stationary transformers: Exploring the stationarity in time series forecasting.
\newblock \emph{Advances in Neural Information Processing Systems}, 35:\penalty0 9881--9893, 2022.

\bibitem[Liu et~al.(2023{\natexlab{b}})Liu, Hu, Zhang, Wu, Wang, Ma, and Long]{liu2023itransformer}
Liu, Y., Hu, T., Zhang, H., Wu, H., Wang, S., Ma, L., and Long, M.
\newblock itransformer: Inverted transformers are effective for time series forecasting.
\newblock \emph{arXiv preprint arXiv:2310.06625}, 2023{\natexlab{b}}.

\bibitem[Lu et~al.(2022)Lu, Grover, Abbeel, and Mordatch]{lu2022frozen}
Lu, K., Grover, A., Abbeel, P., and Mordatch, I.
\newblock Frozen pretrained transformers as universal computation engines.
\newblock In \emph{Proceedings of the AAAI Conference on Artificial Intelligence}, volume~36, pp.\  7628--7636, 2022.

\bibitem[Makridakis et~al.(2018)Makridakis, Spiliotis, and Assimakopoulos]{makridakis2018m4}
Makridakis, S., Spiliotis, E., and Assimakopoulos, V.
\newblock The m4 competition: Results, findings, conclusion and way forward.
\newblock \emph{International Journal of Forecasting}, 34\penalty0 (4):\penalty0 802--808, 2018.

\bibitem[Nate~Gruver \& Wilson(2023)Nate~Gruver and Wilson]{gruver2023llmtime}
Nate~Gruver, Marc~Finzi, S.~Q. and Wilson, A.~G.
\newblock {Large Language Models Are Zero Shot Time Series Forecasters}.
\newblock In \emph{Advances in Neural Information Processing Systems}, 2023.

\bibitem[Nie et~al.(2023)Nie, H.~Nguyen, Sinthong, and Kalagnanam]{Yuqietal-2023-PatchTST}
Nie, Y., H.~Nguyen, N., Sinthong, P., and Kalagnanam, J.
\newblock A time series is worth 64 words: Long-term forecasting with transformers.
\newblock In \emph{International Conference on Learning Representations}, 2023.

\bibitem[Oreshkin et~al.(2019)Oreshkin, Carpov, Chapados, and Bengio]{oreshkin2019n}
Oreshkin, B.~N., Carpov, D., Chapados, N., and Bengio, Y.
\newblock N-beats: Neural basis expansion analysis for interpretable time series forecasting.
\newblock \emph{arXiv preprint arXiv:1905.10437}, 2019.

\bibitem[Ouyang et~al.(2022)Ouyang, Wu, Jiang, Almeida, Wainwright, Mishkin, Zhang, Agarwal, Slama, Ray, et~al.]{ouyang2022training}
Ouyang, L., Wu, J., Jiang, X., Almeida, D., Wainwright, C., Mishkin, P., Zhang, C., Agarwal, S., Slama, K., Ray, A., et~al.
\newblock Training language models to follow instructions with human feedback.
\newblock \emph{Advances in Neural Information Processing Systems}, 35:\penalty0 27730--27744, 2022.

\bibitem[Pan et~al.(2024)Pan, Jiang, Song, Garg, Rasul, Schneider, and Nevmyvaka]{pan2024structural}
Pan, Z., Jiang, Y., Song, D., Garg, S., Rasul, K., Schneider, A., and Nevmyvaka, Y.
\newblock Structural knowledge informed continual multivariate time series forecasting.
\newblock \emph{arXiv preprint arXiv:2402.12722}, 2024.

\bibitem[Paszke et~al.(2019)Paszke, Gross, Massa, Lerer, Bradbury, Chanan, Killeen, Lin, Gimelshein, Antiga, et~al.]{paszke2019pytorch}
Paszke, A., Gross, S., Massa, F., Lerer, A., Bradbury, J., Chanan, G., Killeen, T., Lin, Z., Gimelshein, N., Antiga, L., et~al.
\newblock Pytorch: An imperative style, high-performance deep learning library.
\newblock \emph{Advances in neural information processing systems}, 32, 2019.

\bibitem[Qin et~al.(2017)Qin, Song, Cheng, Cheng, Jiang, and Cottrell]{qin2017dual}
Qin, Y., Song, D., Cheng, H., Cheng, W., Jiang, G., and Cottrell, G.~W.
\newblock A dual-stage attention-based recurrent neural network for time series prediction.
\newblock In \emph{Proceedings of the 26th International Joint Conference on Artificial Intelligence}, pp.\  2627--2633, 2017.

\bibitem[Radford et~al.(2018)Radford, Narasimhan, Salimans, Sutskever, et~al.]{radford2018improving}
Radford, A., Narasimhan, K., Salimans, T., Sutskever, I., et~al.
\newblock Improving language understanding by generative pre-training.
\newblock 2018.

\bibitem[Radford et~al.(2019)Radford, Wu, Child, Luan, Amodei, Sutskever, et~al.]{radford2019language}
Radford, A., Wu, J., Child, R., Luan, D., Amodei, D., Sutskever, I., et~al.
\newblock Language models are unsupervised multitask learners.
\newblock \emph{OpenAI blog}, 1\penalty0 (8):\penalty0 9, 2019.

\bibitem[Raffel et~al.(2020)Raffel, Shazeer, Roberts, Lee, Narang, Matena, Zhou, Li, and Liu]{raffel2020exploring}
Raffel, C., Shazeer, N., Roberts, A., Lee, K., Narang, S., Matena, M., Zhou, Y., Li, W., and Liu, P.~J.
\newblock Exploring the limits of transfer learning with a unified text-to-text transformer.
\newblock \emph{The Journal of Machine Learning Research}, 21\penalty0 (1):\penalty0 5485--5551, 2020.

\bibitem[Rasul et~al.(2023)Rasul, Ashok, Williams, Khorasani, Adamopoulos, Bhagwatkar, Bilo{\v{s}}, Ghonia, Hassen, Schneider, et~al.]{rasul2023lag}
Rasul, K., Ashok, A., Williams, A.~R., Khorasani, A., Adamopoulos, G., Bhagwatkar, R., Bilo{\v{s}}, M., Ghonia, H., Hassen, N.~V., Schneider, A., et~al.
\newblock Lag-llama: Towards foundation models for time series forecasting.
\newblock \emph{arXiv preprint arXiv:2310.08278}, 2023.

\bibitem[Shang et~al.(2021)Shang, Chen, and Bi]{shang2021discrete}
Shang, C., Chen, J., and Bi, J.
\newblock Discrete graph structure learning for forecasting multiple time series.
\newblock \emph{arXiv preprint arXiv:2101.06861}, 2021.

\bibitem[Singhal et~al.(2022)Singhal, Azizi, Tu, Mahdavi, Wei, Chung, Scales, Tanwani, Cole-Lewis, Pfohl, et~al.]{singhal2022large}
Singhal, K., Azizi, S., Tu, T., Mahdavi, S.~S., Wei, J., Chung, H.~W., Scales, N., Tanwani, A., Cole-Lewis, H., Pfohl, S., et~al.
\newblock Large language models encode clinical knowledge.
\newblock \emph{arXiv preprint arXiv:2212.13138}, 2022.

\bibitem[Sun et~al.(2023)Sun, Li, Li, and Hong]{sun2023test}
Sun, C., Li, Y., Li, H., and Hong, S.
\newblock Test: Text prototype aligned embedding to activate llm's ability for time series.
\newblock \emph{arXiv preprint arXiv:2308.08241}, 2023.

\bibitem[Tang et~al.(2022)Tang, Qu, Chow, Lam, Wong, and Ma]{tang2022domain}
Tang, Y., Qu, A., Chow, A.~H., Lam, W.~H., Wong, S., and Ma, W.
\newblock Domain adversarial spatial-temporal network: a transferable framework for short-term traffic forecasting across cities.
\newblock In \emph{Proceedings of the 31st ACM International Conference on Information \& Knowledge Management}, pp.\  1905--1915, 2022.

\bibitem[Taylor \& Letham(2018)Taylor and Letham]{taylor2018forecasting}
Taylor, S.~J. and Letham, B.
\newblock Forecasting at scale.
\newblock \emph{The American Statistician}, 72\penalty0 (1):\penalty0 37--45, 2018.

\bibitem[Tian~Zhou \& Jin(2023)Tian~Zhou and Jin]{zhou2023onefitsall}
Tian~Zhou, Peisong~Niu, X. W. L.~S. and Jin, R.
\newblock {One Fits All}: Power general time series analysis by pretrained lm.
\newblock In \emph{NeurIPS}, 2023.

\bibitem[Touvron et~al.(2023{\natexlab{a}})Touvron, Lavril, Izacard, Martinet, Lachaux, Lacroix, Rozi{\`e}re, Goyal, Hambro, Azhar, et~al.]{touvron2023llama}
Touvron, H., Lavril, T., Izacard, G., Martinet, X., Lachaux, M.-A., Lacroix, T., Rozi{\`e}re, B., Goyal, N., Hambro, E., Azhar, F., et~al.
\newblock Llama: Open and efficient foundation language models.
\newblock \emph{arXiv preprint arXiv:2302.13971}, 2023{\natexlab{a}}.

\bibitem[Touvron et~al.(2023{\natexlab{b}})Touvron, Lavril, Izacard, Martinet, Lachaux, Lacroix, Rozi{\`e}re, Goyal, Hambro, Azhar, et~al.]{touvron2023llama1}
Touvron, H., Lavril, T., Izacard, G., Martinet, X., Lachaux, M.-A., Lacroix, T., Rozi{\`e}re, B., Goyal, N., Hambro, E., Azhar, F., et~al.
\newblock Llama: Open and efficient foundation language models.
\newblock \emph{arXiv preprint arXiv:2302.13971}, 2023{\natexlab{b}}.

\bibitem[Touvron et~al.(2023{\natexlab{c}})Touvron, Martin, Stone, Albert, Almahairi, Babaei, Bashlykov, Batra, Bhargava, Bhosale, et~al.]{touvron2023llama2}
Touvron, H., Martin, L., Stone, K., Albert, P., Almahairi, A., Babaei, Y., Bashlykov, N., Batra, S., Bhargava, P., Bhosale, S., et~al.
\newblock Llama 2: Open foundation and fine-tuned chat models.
\newblock \emph{arXiv preprint arXiv:2307.09288}, 2023{\natexlab{c}}.

\bibitem[Woo et~al.(2022)Woo, Liu, Sahoo, Kumar, and Hoi]{woo2022etsformer}
Woo, G., Liu, C., Sahoo, D., Kumar, A., and Hoi, S.
\newblock Etsformer: Exponential smoothing transformers for time-series forecasting.
\newblock \emph{arXiv preprint arXiv:2202.01381}, 2022.

\bibitem[Wu et~al.(2021)Wu, Xu, Wang, and Long]{wu2021autoformer}
Wu, H., Xu, J., Wang, J., and Long, M.
\newblock Autoformer: Decomposition transformers with auto-correlation for long-term series forecasting.
\newblock \emph{Advances in Neural Information Processing Systems}, 34:\penalty0 22419--22430, 2021.

\bibitem[Wu et~al.(2023)Wu, Hu, Liu, Zhou, Wang, and Long]{wu2023timesnet}
Wu, H., Hu, T., Liu, Y., Zhou, H., Wang, J., and Long, M.
\newblock Timesnet: Temporal 2d-variation modeling for general time series analysis.
\newblock In \emph{International Conference on Learning Representations}, 2023.

\bibitem[Wu et~al.(2020)Wu, Pan, Long, Jiang, Chang, and Zhang]{wu2020connecting}
Wu, Z., Pan, S., Long, G., Jiang, J., Chang, X., and Zhang, C.
\newblock Connecting the dots: Multivariate time series forecasting with graph neural networks.
\newblock In \emph{Proceedings of the 26th ACM SIGKDD international conference on knowledge discovery \& data mining}, pp.\  753--763, 2020.

\bibitem[Xue \& Salim(2022)Xue and Salim]{xue2022promptcast}
Xue, H. and Salim, F.~D.
\newblock Promptcast: A new prompt-based learning paradigm for time series forecasting.
\newblock 2022.

\bibitem[Yin et~al.(2023)Yin, Fu, Zhao, Li, Sun, Xu, and Chen]{yin2023survey}
Yin, S., Fu, C., Zhao, S., Li, K., Sun, X., Xu, T., and Chen, E.
\newblock A survey on multimodal large language models.
\newblock \emph{arXiv preprint arXiv:2306.13549}, 2023.

\bibitem[Zeng et~al.(2023)Zeng, Chen, Zhang, and Xu]{zeng2023transformers}
Zeng, A., Chen, M., Zhang, L., and Xu, Q.
\newblock Are transformers effective for time series forecasting?
\newblock In \emph{Proceedings of the AAAI conference on artificial intelligence}, volume~37, pp.\  11121--11128, 2023.

\bibitem[Zhang et~al.(2022{\natexlab{a}})Zhang, Zhang, Cao, Bian, Yi, Zheng, and Li]{zhang2022less}
Zhang, T., Zhang, Y., Cao, W., Bian, J., Yi, X., Zheng, S., and Li, J.
\newblock Less is more: Fast multivariate time series forecasting with light sampling-oriented mlp structures.
\newblock \emph{arXiv preprint arXiv:2207.01186}, 2022{\natexlab{a}}.

\bibitem[Zhang et~al.(2022{\natexlab{b}})Zhang, Zhao, Tsiligkaridis, and Zitnik]{zhang2022self}
Zhang, X., Zhao, Z., Tsiligkaridis, T., and Zitnik, M.
\newblock Self-supervised contrastive pre-training for time series via time-frequency consistency.
\newblock \emph{Advances in Neural Information Processing Systems}, 35:\penalty0 3988--4003, 2022{\natexlab{b}}.

\bibitem[Zhou et~al.(2021)Zhou, Zhang, Peng, Zhang, Li, Xiong, and Zhang]{zhou2021informer}
Zhou, H., Zhang, S., Peng, J., Zhang, S., Li, J., Xiong, H., and Zhang, W.
\newblock Informer: Beyond efficient transformer for long sequence time-series forecasting.
\newblock In \emph{Proceedings of the AAAI conference on artificial intelligence}, volume~35, pp.\  11106--11115, 2021.

\bibitem[Zhou et~al.(2022)Zhou, Ma, Wen, Wang, Sun, and Jin]{zhou2022fedformer}
Zhou, T., Ma, Z., Wen, Q., Wang, X., Sun, L., and Jin, R.
\newblock Fedformer: Frequency enhanced decomposed transformer for long-term series forecasting.
\newblock In \emph{International Conference on Machine Learning}, pp.\  27268--27286. PMLR, 2022.

\end{thebibliography}

\bibliographystyle{icml2024}

\newpage
\appendix
\onecolumn

\section{Experimental Details}\label{greedy}

\subsection{Implementation}\label{A1}

We mainly follow the experimental configurations in~\citep{wu2023timesnet} across all baselines within a unified evaluation pipeline, available at https://github.com/thuml/Time-Series-Library, for a fair comparison. We use GPT2-small~\citep{radford2019language} with the first 6 hidden layers enabled as the default backbone model.  All our experiments are repeated three times and we report the averaged results. We implemented the model on PyTorch~\cite {paszke2019pytorch} with all experiments conducted on NVIDIA RTX A6000 GPUs \textcolor{black}{and NVIDIA A100 GPUs. We configure the patch length P as 16 with a stride S of 8. Our experimental results were obtained by performing a grid search over combinations within the following search spaces: long-term trend length (24, 48, 96, 144), seasonal trend length (2, 4, 8, 12, 24), prompt length (2, 4, 8, 16, 32), and semantic space size (1000, 2000, 5000).}  \textcolor{black}{Our code is available at \href{https://github.com/panzijie825/S2IP-LLM}{https://github.com/panzijie825/S2IP-LLM.}}

\subsection{Baseline Introduction}\label{baseline_intro}
We introduce the baseline models that we choose to compare in the following section:

\begin{itemize}[noitemsep,topsep=0pt,parsep=0pt,partopsep=0pt]
    \item \textbf{Time-LLM}~\citep{jin2023time}: Time-LLM reprograms time series tokens with NLP representation using multi-head attention and fine-tunes the pre-trained LLM with the prefix prompting to perform time series analysis. \textcolor{black}{We reproduced and reported the experimental results for long-term forecasting using two LLM backbones: `L' represents LLaMA~\citep{touvron2023llama}, and `G' represents GPT-2~\citep{radford2019language}.}
    \item \textbf{OFA}~\citep{zhou2023onefitsall}: OFA represents time series data into patched tokens to fine-tune the pre-trained GPT2~\citep{radford2019language} for various time series analysis tasks. 
    \item \textbf{iTransformer}~\citep{liu2023itransformer}: iTransformer applies the attention and feed-forward network on the inverted dimensions of the time series data to capture multivariate correlations.
    \item \textbf{Dlinear}~\citep{zeng2023transformers}: Dlinear incorporates the decomposition with linear layer to model the time series data via modeling trend and seasonal components separately.
    \item \textbf{PatchTST}~\citep{Yuqietal-2023-PatchTST}: PatchTST leverages a Transformer-based model for time series forecasting by segmenting data into patches and using a channel-independent design to efficiently reduce computational costs and boost forecasting performance.
    \item \textbf{TimesNet}~\citep{wu2023timesnet}: TimesNet converts 1D time series data into 2D representation and capture intra- and inter-period relations. It designs TimesBlock with an inception block to extract complex temporal patterns, leading to multiple time series tasks.
    \item \textbf{FEDformer}~\citep{zhou2022fedformer}: FEDformer incorporates seasonal-trend decomposition with Transformers for time series forecasting. It leverages information from the frequency domain, gaining efficiency and accuracy in time series analysis. 
    \item \textbf{Autoformer}~\citep{wu2021autoformer}: Autoformer proposes the decomposition architecture with Auto-Correlation mechanisms to efficiently and accurately perform long-term forecasting.
    \item \textbf{Stationary}~\citep{liu2022non}: Non-stationary Transformers proposes a framework with two interdependent modules, namely series stationarization and de-stationary attention to gain robust time series forecasting results.
    \item \textbf{ETSformer}~\citep{woo2022etsformer}: ETSformer integrates exponential smoothing principles by replacing traditional self-attention with exponential smoothing attention and frequency attention for time series forecasting 
    \item \textbf{LightTS}~\citep{zhang2022less}: LightTS is a time series classification framework that includes adaptive ensemble distillation and Pareto optimization, resulting in accurate classification with limited resources.
\end{itemize}

We note that patching-based methods, \textit{i.e.} OFA~\citep{zhou2023onefitsall}, PatchTST~\citep{Yuqietal-2023-PatchTST}, and Time-LLM~\citep{jin2023time} treat multivariate time series as independently univariate time series, which essentially provide more training data for those models. For transformer-based models which rely on multivariate times input, this could be the reason that their performances are not as good as patching-based ones.

\subsection{Details of Datasets}\label{A3}

We experiment the long-term forecasting on the widely adopted Electricity Transformer Temperature (ETT) datasets~\citep{zhou2021informer}, Weather, Electricity, and Traffic from~\citep{wu2023timesnet}. We also experiment the short-term forecasting using the M4 benchmark dataset~\citep{makridakis2018m4}.\\
ETT datasets are comprised of roughly two years of data from two locations in China. The data are further divided into four distinct datasets, each with different sampling rates: ETTh1 and ETTh2 are sampled hourly, and ETTm1 and ETTm2 are sampled every 15 minutes. Every ETT dataset includes six power load features and a target variable: the oil temperature. The Electricity dataset comprises records of electricity consumption from 321 customers and is measured with a 1-hour sampling rate. The Weather dataset contains one-year records from 21 meteorological stations located in Germany. The sampling rate for the Weather dataset is 10 minutes. The Traffic dataset includes the per-hour sampled occupancy rates of the freeway system, which were recorded from 862 sensors in California. The M4 benchmark dataset has 100 thousand time series, which were collected from various domains ranging from business to economic forecasting. The time series data are partitioned into six groups with varied sampling rates from yearly to hourly. 

The full data statistics are summarized in Table~\ref{data_stats}

\begin{table*}[!h]
\caption{Dataset statistics are from \citep{wu2023timesnet}. The dimension indicates the number of time series variables, and the dataset size is organized in (training, validation, and testing).}

\begin{tabular}{cl|cl|c|cl|cll|cl|cl}
\toprule
\cmidrule{1-14}
\multicolumn{2}{c|}{Tasks} &
  \multicolumn{2}{c|}{Datasets} &
  Dim. &
  \multicolumn{2}{c|}{Series Length} &
  \multicolumn{3}{c|}{Dataset Size} &
  \multicolumn{2}{c|}{Frequency} &
  \multicolumn{2}{c}{Information} \\ 
  \midrule
  \cmidrule{1-14}
\multicolumn{2}{c|}{\multirow{4}{*}{\begin{tabular}[c]{@{}c@{}}Long-term \\ Forecasting\end{tabular}}} &
  \multicolumn{2}{c|}{ETTm1} &
  7 &
  \multicolumn{2}{c|}{\{96,192,336,720\}} &
  \multicolumn{3}{c|}{(34465, 11521, 11521)} &
  \multicolumn{2}{c|}{15 min} &
  \multicolumn{2}{c}{Temperature} \\
\multicolumn{2}{c|}{} &
  \multicolumn{2}{c|}{ETTm2} &
  7 &
  \multicolumn{2}{c|}{\{96,192,336,720\}} &
  \multicolumn{3}{c|}{(34465, 11521, 11521)} &
  \multicolumn{2}{c|}{15 min} &
  \multicolumn{2}{c}{Temperature} \\
\multicolumn{2}{c|}{} &
  \multicolumn{2}{c|}{ETTh1} &
  7 &
  \multicolumn{2}{c|}{\{96,192,336,720\}} &
  \multicolumn{3}{c|}{(8545, 2881, 2881)} &
  \multicolumn{2}{c|}{1 hour} &
  \multicolumn{2}{c}{Temperature} \\
\multicolumn{2}{c|}{} &
  \multicolumn{2}{c|}{ETTh2} &
  7 &
  \multicolumn{2}{c|}{\{96,192,336,720\}} &
  \multicolumn{3}{c|}{(8545, 2881, 2881)} &
  \multicolumn{2}{c|}{1 hour} &
  \multicolumn{2}{c}{Temperature} \\
\multicolumn{2}{c|}{} &
  \multicolumn{2}{c|}{Electricity} &
  321 &
  \multicolumn{2}{c|}{\{96,192,336,720\}} &
  \multicolumn{3}{c|}{(18317, 2633, 5261)} &
  \multicolumn{2}{c|}{1 hour} &
  \multicolumn{2}{c}{Electricity} \\
\multicolumn{2}{c|}{} &
  \multicolumn{2}{c|}{Traffic} &
  862 &
  \multicolumn{2}{c|}{\{96,192,336,720\}} &
  \multicolumn{3}{c|}{(12185, 1757, 3509)} &
  \multicolumn{2}{c|}{1 hour} &
  \multicolumn{2}{c}{Transportation} \\
\multicolumn{2}{c|}{} &
  \multicolumn{2}{c|}{Weather} &
  21 &
  \multicolumn{2}{c|}{\{96,192,336,720\}} &
  \multicolumn{3}{c|}{(36792, 5271, 10540)} &
  \multicolumn{2}{c|}{10 min} &
  \multicolumn{2}{c}{Weather} \\
  
\midrule
\cmidrule{1-14}
\multicolumn{2}{c|}{\multirow{4}{*}{\begin{tabular}[c]{@{}c@{}}Short-term \\ Forecasting\end{tabular}}} &
  \multicolumn{2}{c|}{M4-Yearly} &
   1 &
  \multicolumn{2}{c|}{6} &
  \multicolumn{3}{c|}{(23000, 0, 23000)} &
  \multicolumn{2}{c|}{Yearly} &
  \multicolumn{2}{c}{Demographic} \\
\multicolumn{2}{c|}{} &
  \multicolumn{2}{c|}{M4-Quarterly} &
  1 &
  \multicolumn{2}{c|}{8} &
  \multicolumn{3}{c|}{(24000, 0, 24000)} &
  \multicolumn{2}{c|}{Quarterly} &
  \multicolumn{2}{c}{Finance} \\
\multicolumn{2}{c|}{} &
  \multicolumn{2}{c|}{M4-Monthly} &
  1 &
  \multicolumn{2}{c|}{18} &
  \multicolumn{3}{c|}{(48000, 0, 48000)} &
  \multicolumn{2}{c|}{Monthly} &
  \multicolumn{2}{c}{Industry} \\
\multicolumn{2}{c|}{} &
  \multicolumn{2}{c|}{M4-Weekly} &
  1 &
  \multicolumn{2}{c|}{13} &
  \multicolumn{3}{c|}{(359, 0, 359)} &
  \multicolumn{2}{c|}{Weekly} &
  \multicolumn{2}{c}{Macro}\\
\multicolumn{2}{c|}{} &
\multicolumn{2}{c|}{M4-Daily} &
  1 &
  \multicolumn{2}{c|}{14} &
  \multicolumn{3}{c|}{(4227, 0, 4227)} &
  \multicolumn{2}{c|}{Daily} &
  \multicolumn{2}{c}{Micro}\\
\multicolumn{2}{c|}{} &
  \multicolumn{2}{c|}{M4-Hourly} &
  1 &
  \multicolumn{2}{c|}{48} &
  \multicolumn{3}{c|}{(414, 0, 414)} &
  \multicolumn{2}{c|}{Hourly} &
  \multicolumn{2}{c}{Other}\\
  
\bottomrule
\end{tabular}
\label{data_stats}
\end{table*}

\subsection{Evaluation Metrics}\label{A4}

For evaluation metrics, we use the mean square error (MSE) and mean absolute error (MAE) for long-term forecasting. For short-term forecasting on the M4 benchmark, we use the symmetric mean absolute percentage error (SMAPE), mean absolute scaled error (MASE), and overall weighted average (OWA)~\citep{oreshkin2019n}, which is a specific metric
for the M4 competition. We present the calculations of these metrics as follows:

$\begin{aligned} & \text { MSE }=\frac{1}{H} \sum_{h=1}^T\left(\mathbf{Y}_h-\hat{\mathbf{Y}}_h\right)^2, \quad \quad \text { MAE }=\frac{1}{H} \sum_{h=1}^H\left|\mathbf{Y}_h-\hat{\mathbf{Y}}_h\right|, \\ & \text { SMAPE }=\frac{200}{H} \sum_{h=1}^H \frac{\left|\mathbf{Y}_h-\hat{\mathbf{Y}}_h\right|}{\left|\mathbf{Y}_h\right|+\left|\hat{\mathbf{Y}}_h\right|}, \quad \text { MAPE }=\frac{100}{H} \sum_{h=1}^H \frac{\left|\mathbf{Y}_h-\hat{\mathbf{Y}}_h\right|}{\left|\mathbf{Y}_h\right|}, \\ & \text { MASE }=\frac{1}{H} \sum_{h=1}^H \frac{\left|\mathbf{Y}_h-\hat{\mathbf{Y}}_h\right|}{\frac{1}{H-s} \sum_{j=s+1}^H\left|\mathbf{Y}_j-\mathbf{Y}_{j-s}\right|}, \quad \text { OWA }=\frac{1}{2}\left[\frac{\text { SMAPE }}{\text { SMAPE }_{\text {Naïve2 }}}+\frac{\text { MASE }}{\text { MASE }_{\text {Naïve2 }}}\right] \\ & \end{aligned}$

where s is the time series data periodicity. H denotes the prediction intervals. $Y_h$ and $\hat{Y_h}$ are the $h$-th ground truth and prediction where h $\in \{1,..., H\}$. For the evaluation metrics in long-term forecasting, we clarify that the reported metrics are the normalized versions of MAE/MSE. Although we apply global standardization to the data, the information that the scaler used is from training data solely.

\section{Long-term Forecasting Results}\label{sec:appendix:long-term}

Table~\ref{tab:long-term-forecast-transformer} shows the detailed results of all prediction lengths of five Transformer-based forecasting models. $S^2$IP-LLM shows a strong and relatively stable performance across different datasets compared to other transformer-based models. 


We note that patching-based methods, \textit{i.e.} OFA~\citep{zhou2023onefitsall}, PatchTST~\citep{Yuqietal-2023-PatchTST} treat multivariate time series independently as univariate time series, which essentially provide more training data for univariate time series input based models. This potentially contributes to their advantages in terms of performance. Thus, it may create an unfair comparison to methods with truly multivariate inputs, \textit{i.e.} the other transformer-based models.

\begin{table*}[t]
\caption{{Transformer-based Models Long-term forecasting results for \{96, 192, 336, 720\} horizons. A lower value indicates a better performance.}}
\scriptsize
\centering
\begin{tabular}{ll|ll|ll|ll|ll|ll|ll|ll}
\hline
\multicolumn{2}{c|}{Methods} &
  \multicolumn{2}{c|}{\textbf{$\mathbf{S^2}$IP-LLM}} &
  \multicolumn{2}{c|}{\textbf{TimesNet}} &
  \multicolumn{2}{c|}{\textbf{FEDformer}} &
  \multicolumn{2}{c|}{\textbf{Autoformer}} &
  \multicolumn{2}{c|}{\textbf{Stationary}} &
  \multicolumn{2}{c|}{\textbf{ETSformer}} &
  \multicolumn{2}{c}{\textbf{LightTS}}   \\ \hline
\multicolumn{2}{c|}{$\text{Datasets} \backslash \text{Horizon}$} &
  \multicolumn{1}{c}{MSE} &
  \multicolumn{1}{c|}{MAE} &
  \multicolumn{1}{c}{MSE} &
  \multicolumn{1}{c|}{MAE} &
  \multicolumn{1}{c}{MSE} &
  \multicolumn{1}{c|}{MAE} &
  \multicolumn{1}{c}{MSE} &
  \multicolumn{1}{c|}{MAE} &
  \multicolumn{1}{c}{MSE} &
  \multicolumn{1}{c|}{MAE} &
  \multicolumn{1}{c}{MSE} &
  \multicolumn{1}{c|}{MAE}&
  \multicolumn{1}{c}{MSE} &
  \multicolumn{1}{c}{MAE}\\ \hline

\multicolumn{1}{l|}{\multirow{5}{*}{\textbf{Weather}}} 
& 96 &  \textbf{0.145} & \textbf{0.195} & 0.172 & 0.220 & 0.217 & 0.296 & 0.266 & 0.336 & 0.173 & 0.223 & 0.197 & 0.281 & 0.182 & 0.242 \\
\multicolumn{1}{l|}{}                       
& 192 & \textbf{\textcolor{black}{0.190}} & \textbf{\textcolor{black}{0.235}}&0.219 & 0.261& 0.276 & 0.336 & 0.307 & 0.367 & 0.245 & 0.285 & 0.237 & 0.312 & 0.227 & 0.287  \\
\multicolumn{1}{l|}{}                       
& 336 & \textbf{\textcolor{black}{0.243}} & \textbf{\textcolor{black}{0.280}} & 0.280 & 0.306 & 0.339 & 0.380 & 0.359 & 0.395 & 0.321 & 0.338 & 0.298 & 0.353 & 0.282 & 0.334  \\ 
\multicolumn{1}{l|}{}                      
& 720 & \textbf{\textcolor{black}{0.312}} & \textbf{\textcolor{black}{0.326}} & 0.365 & 0.359 & 0.403 & 0.428 & 0.419 & 0.428 & 0.414 & 0.410 & 0.352 & 0.288 & 0.352 & 0.386
\\
\multicolumn{1}{l|}{}                       
& Avg & \textbf{\textcolor{black}{0.222}} & \textbf{\textcolor{black}{0.259}} & 0.259 & 0.287 & 0.309 & 0.360 & 0.338 & 0.382 & 0.288 & 0.314 & 0.271 & 0.334 & 0.261 & 0.312  \\

\hline

\multicolumn{1}{l|}{\multirow{5}{*}{\textbf{Electricity}}} 
& 96  & \underline{\textcolor{black}{0.135}} & \underline{\textcolor{black}{0.230}} & 0.168 & 0.272  & 0.193 & 0.308 & 0.201 & 0.317 & 0.169 & 0.273 & 0.187 & 0.304 & 0.207 & 0.307 \\
\multicolumn{1}{l|}{}                       
& 192 & \textbf{\textcolor{black}{0.149}} & \underline{\textcolor{black}{0.247}} & 0.184 & 0.289 & 0.201 & 0.315 & 0.222 & 0.334 & 0.182 & 0.286 & 0.199 & 0.315 & 0.213 & 0.316   \\
\multicolumn{1}{l|}{}                       
& 336 & \underline{\textcolor{black}{0.167}} & \underline{\textcolor{black}{0.266}} & 0.198 & 0.300 & 0.214 & 0.329 & 0.231 & 0.338 & 0.200 & 0.304 & 0.212 & 0.329 & 0.230 & 0.333  \\ 
\multicolumn{1}{l|}{}                      
& 720 & \underline{\textcolor{black}{0.200}} & \textbf{\textcolor{black}{0.287}} & 0.220 & 0.320  & 0.246 & 0.355 & 0.254 & 0.361 & 0.222 & 0.321 & 0.233 & 0.345 & 0.265 & 0.360  \\
\multicolumn{1}{l|}{}                       
& Avg & \textbf{\textcolor{black}{0.161}} & \underline{\textcolor{black}{0.257}} & 0.192 & 0.295 & 0.214 & 0.327 & 0.227 & 0.338 & 0.193 & 0.296 & 0.208 & 0.323 & 0.229 & 0.329  \\

\hline

\multicolumn{1}{l|}{\multirow{5}{*}{\textbf{Traffic}}} 
& 96  & \textcolor{black}{0.379} & \textcolor{black}{0.274} & 0.593 & 0.321  & 0.587 & 0.366 & 0.613 & 0.388 & 0.612 & 0.338 & 0.607 & 0.392 & 0.615 & 0.391 \\
\multicolumn{1}{l|}{}                       
& 192 & \textcolor{black}{0.397} & \underline{\textcolor{black}{0.282}} & 0.617 & 0.336  & 0.604 & 0.373 & 0.616 & 0.382 & 0.613 & 0.340 & 0.621 & 0.399 & 0.601 & 0.382   \\
\multicolumn{1}{l|}{}                       
& 336 & \textcolor{black}{0.407} & \underline{\textcolor{black}{0.289}} & 0.629 & 0.336  & 0.621 & 0.383 & 0.622 & 0.337 & 0.618 & 0.328 & 0.622 & 0.396 & 0.613 & 0.386  \\ 
\multicolumn{1}{l|}{}                      
& 720& \textcolor{black}{0.440} & \underline{\textcolor{black}{0.301}} & 0.640 & 0.350 & 0.626 & 0.382 & 0.660 & 0.408 & 0.653 & 0.355 & 0.632 & 0.396 & 0.658 & 0.407 \\
\multicolumn{1}{l|}{}                       
& Avg & \textcolor{black}{0.405} & \underline{\textcolor{black}{0.286}} & 0.620 & 0.336 & 0.610 & 0.376 & 0.628 & 0.379 & 0.624 & 0.340 & 0.621 & 0.396 & 0.622 & 0.392  \\

\hline

\multicolumn{1}{l|}{\multirow{5}{*}{\textbf{ETTh1}}} 

& 96 & \textbf{\textcolor{black}{0.366}} & \textbf{\textcolor{black}{0.396}} & 0.468 & 0.475 & 0.376 & 0.419 & 0.530 & 0.517 & 0.513 & 0.491 & 0.644 & 0.589& 0.440 & 0.450 \\
\multicolumn{1}{l|}{}                       
& 192 & \textbf{\textcolor{black}{0.401}} & \underline{\textcolor{black}{0.420}} & 0.484 & 0.485 & 0.420 & 0.448 & 0.537 & 0.521 & 0.534 & 0.504 & 0.736 & 0.648 & 0.498 & 0.479  \\
\multicolumn{1}{l|}{}                       
& 336 & \textbf{\textcolor{black}{0.412}} & \textbf{\textcolor{black}{0.431}} & 0.536 & 0.516 & 0.459 & 0.465 & 0.596 & 0.583 & 0.588 & 0.535 & 0.827 & 0.707 & 0.550 & 0.510  \\
\multicolumn{1}{l|}{}                      
& 720 & \underline{\textcolor{black}{0.440}} & \underline{\textcolor{black}{0.458}}& 0.593 & 0.537 & 0.506 & 0.507 & 0.713 & 0.639 & 0.643 & 0.616 & 0.946 & 0.766 & 0.615 & 0.571   \\
\multicolumn{1}{l|}{}                       
& Avg & \textbf{\textcolor{black}{0.406}} & \textbf{\textcolor{black}{0.427}} & 0.520 & 0.505 & 0.440 &  0.460 & 0.594 & 0.565 & 0.570 & 0.537 & 0.788 & 0.677 & 0.526 & 0.502  \\ 

\hline

\multicolumn{1}{l|}{\multirow{5}{*}{\textbf{ETTh2}}} 

& 96 & \textbf{\textcolor{black}{0.278}} & \textbf{\textcolor{black}{0.340}} & 0.376 & 0.415 & 0.358 & 0.397 & 0.454 & 0.490 & 0.476 & 0.458 & 0.340 & 0.391  & 0.408 & 0.445  \\
\multicolumn{1}{l|}{}                       
& 192 & \textbf{\textcolor{black}{0.346}} & \textbf{\textcolor{black}{0.385}} & 0.409 & 0.440  & 0.429 & 0.439 & 0.486 & 0.517 & 0.512 & 0.493 & 0.430 & 0.439 & 0.561 & 0.526   \\
\multicolumn{1}{l|}{}                       
& 336 & \textbf{\textcolor{black}{0.367}} & \textbf{\textcolor{black}{0.406}} & 0.425 & 0.455 & 0.496 & 0.487 & 0.493 & 0.533 & 0.552 & 0.551 & 0.485 & 0.479 & 0.673 & 0.580 \\
\multicolumn{1}{l|}{}                      
& 720 & \underline{\textcolor{black}{0.400}} & \underline{\textcolor{black}{0.436}} & 0.488 & 0.494 & 0.463 & 0.474 & 0.515 & 0.543 & 0.562 & 0.560 & 0.500 & 0.497 & 1.006 & 0.721  \\
\multicolumn{1}{l|}{}                       
& Avg & \textbf{\textcolor{black}{0.347}} & \textbf{\textcolor{black}{0.391}}
 & 0.425 & 0.451 & 0.437 & 0.449 & 0.487 & 0.520 & 0.526 & 0.516 &  0.439 & 0.452 & 0.662 & 0.568  \\ 
\hline
\multicolumn{1}{l|}{\multirow{5}{*}{\textbf{ETTm1}}} 

& 96  & \textbf{\textcolor{black}{0.288}} & \textbf{\textcolor{black}{0.346}} & 0.329 & 0.377 & 0.379 & 0.419 & 0.568 & 0.516 & 0.386 & 0.398 & 0.375 & 0.398 & 0.383 & 0.409  \\
\multicolumn{1}{l|}{}                       
& 192 & \textbf{\textcolor{black}{0.323}} & \textbf{\textcolor{black}{0.365}} & 0.371 & 0.401 & 0.426 & 0.441 & 0.573 & 0.528 & 0.459 & 0.444 &  0.408 & 0.410 & 0.421 & 0.431  \\
\multicolumn{1}{l|}{}                       
& 336 & \textbf{\textcolor{black}{0.359}} & \underline{\textcolor{black}{0.390}} & 0.417 & 0.428  & 0.445 & 0.459 & 0.587 & 0.534 & 0.495 & 0.464 & 0.435 & 0.428 & 0.454 & 0.456  \\
\multicolumn{1}{l|}{}                      
& 720 &  \textbf{\textcolor{black}{0.403}} & \textbf{\textcolor{black}{0.418}} & 0.483 & 0.464 & 0.543 & 0.490 & 0.589 & 0.536  & 0.585 &  0.516 & 0.499 & 0.462 & 0.549 & 0.520 \\
\multicolumn{1}{l|}{}                       
& Avg & \textbf{\textcolor{black}{0.343}} & \textbf{\textcolor{black}{0.379}} & 0.400 & 0.417 & 0.448 & 0.452 & 0.579 & 0.529 & 0.481 & 0.456 & 0.429 & 0.425 & 0.452 & 0.454 \\ 
\hline
\multicolumn{1}{l|}{\multirow{5}{*}{\textbf{ETTm2}}} 

& 96  & \textbf{\textcolor{black}{0.165}} & \textbf{\textcolor{black}{0.257}} & 0.201 & 0.286 & 0.203 & 0.287 & 0.287 & 0.359 & 0.192 & 0.274 &  0.189 & 0.280  & 0.239 & 0.335 \\
\multicolumn{1}{l|}{}                       
& 192 & \textbf{\textcolor{black}{0.222}} & \textbf{\textcolor{black}{0.299}} & 0.260 & 0.329  & 0.269 & 0.328 & 0.325 & 0.388 & 0.280 & 0.339 & 0.253 & 0.319 & 0.346 & 0.412  \\
\multicolumn{1}{l|}{}                       
& 336 & \textbf{\textcolor{black}{0.277}} & \textbf{\textcolor{black}{0.330}}& 0.331 & 0.376 & 0.325 & 0.366 & 0.498 & 0.491 & 0.334 & 0.361 & 0.314 & 0.357 & 0.506 & 0.506  \\ 
\multicolumn{1}{l|}{}                      
& 720 &  \textbf{\textcolor{black}{0.363}} & \textbf{\textcolor{black}{0.390}} & 0.428 & 0.430 & 0.421 & 0.415 & 0.548 & 0.517 & 0.417 & 0.413 & 0.414 & 0.413 & 0.702 & 0.606 \\
\multicolumn{1}{l|}{}                       
& Avg & \textbf{\textcolor{black}{0.257}} & \textbf{\textcolor{black}{0.319}} & 0.305 & 0.355 & 0.305 & 0.349 & 0.414 & 0.439 & 0.306 & 0.347 & 0.293 & 0.342 & 0.448 & 0.465  \\

\bottomrule

\end{tabular}
\label{tab:long-term-forecast-transformer}
\end{table*}

\section{Full Short-term Forecasting Results}\label{appendix:short-term}

Table~\ref{tab:short-term-forecast-full} shows the full short-term forecasting experiment results on M4 datasets. $S^2$IP-LLM consistently outperforms the majority of baseline models in most cases. It surpasses the performance of OFA significantly and achieves slightly better forecasting performance than PatchTST, which can be attributed to proposed semantic space informed prompting.

\begin{table*}[!h]
\scriptsize
\setlength{\tabcolsep}{1.2pt}
\centering
\caption{Detailed short-term time series forecasting results on M4 datasets. The forecasting horizons are in [6, 48] and the last three rows are weighted averaged from all datasets under different sampling intervals. A lower value indicates
better performance. } 
\begin{tabular}{cc|cccccccccccccc}
\hline
\multicolumn{2}{c|}{Methods} & \textbf{$\mathbf{S^2}$IP-LLM} & \textbf{Time-LLM(G)} & \textbf{OFA} & \textbf{iTrans.} & \textbf{Dlinear} & \textbf{PatchTST} & \textbf{N-HiTS} & \textbf{N-BEATS} & \textbf{TimesNet} & \textbf{FEDformer} & \textbf{Autoformer} & \textbf{Stationary} & \textbf{ETSformer}  \\ 
\hline

\multirow{3}{*}{Year.} & SMAPE & \textbf{\textcolor{black}{13.413}} & \textcolor{black}{13.75} & 15.11 & 13.652 & 16.965 & 13.477 & \underline{13.422} & 13.487 & 15.378 & 14.021 & 13.974 & 14.727 & 18.009  \\ 
& MASE & \underline{\textcolor{black}{3.024}} & \textcolor{black}{3.055} & 3.565 & 3.095 & 4.283 & \textbf{3.019} & 3.056 & 3.036 & 3.554 & 3.036 & 3.134 & 3.078 & 4.487 \\ 
& OWA & \textbf{\textcolor{black}{0.792}} & \textcolor{black}{0.805} & 0.911 & 0.807 & 1.058 & \textbf{0.792} & \underline{0.795} & \underline{0.795} & 0.918 & 0.811 & 0.822 & 0.807 & 1.115  \\ 
\hline

\multirow{3}{*}{Quart.} & SMAPE & \underline{\textcolor{black}{10.352}} & \textcolor{black}{10.671} & 10.597 & 10.353 & 12.145 & 10.38 & \textbf{10.185} & 10.564 & 10.465 & 11.100 & 11.338 & 10.958 & 13.376  \\  
& MASE& \textcolor{black}{1.228} & \textcolor{black}{1.276} & 1.253 & 1.209 & 1.520 & 1.233 & \textbf{1.18} & 1.252 & \underline{1.227} & 1.35 & 1.365 & 1.325 & 1.906  \\  
& OWA& \textcolor{black}{0.922} & \textcolor{black}{0.95}  & 0.938 & \underline{0.911} & 1.106 & 0.921 & \textbf{0.893} & 0.936 & 0.923 & 0.996 & 1.012 & 0.981 & 1.302  \\ 
\hline

\multirow{3}{*}{Month.} & SMAPE & \underline{\textcolor{black}{12.995}} & \textcolor{black}{13.416} & 13.258 & 13.079 & 13.514 & \textbf{12.959} & 13.059 & 13.089 & 13.513 & 14.403 & 13.958 & 13.917 & 14.588  \\ 
& MASE& \textbf{\textcolor{black}{0.97}} & \textcolor{black}{1.045} & 1.003 & \underline{0.974} & 1.037 & \textbf{0.970} & 1.013 & 0.996 & 1.039 & 1.147 & 1.103 & 1.097 & 1.368  \\ 
& OWA& \underline{\textcolor{black}{0.91}} & \textcolor{black}{0.957} & 0.931 & 0.911 & 0.956 & \textbf{0.905} & 0.929 & 0.922 & 0.957 & 1.038 & 1.002 & 0.998 & 1.149  \\ 
\hline

\multirow{3}{*}{Others.} & SMAPE & \textcolor{black}{4.805} & \textcolor{black}{4.973} & 6.124 & \underline{4.78} & 6.709 & 4.952 & \textbf{4.711} & 6.599 & 6.913 & 7.148 & 5.485 & 6.302 & 7.267  \\  
& MASE & \textcolor{black}{3.247} & \textcolor{black}{3.412} & 4.116 & \underline{3.231} & 4.953 & 3.347 & \textbf{3.054} & 4.430 & 4.507 & 4.064 & 3.865 & 4.064 & 5.240  \\  
& OWA& \textcolor{black}{1.017} & \textcolor{black}{1.053} & 1.259 & \underline{1.012} & 1.487 & 1.049 & \textbf{0.977} & 1.393 & 1.438 & 1.304 & 1.187 & 1.304 & 1.591\\ 
\hline

\multirow{3}{*}{Avg.} & SMAPE & \textbf{\textcolor{black}{12.021}} & \textcolor{black}{12.494} & 12.690 & 12.142 & 13.639 & 12.059 & \underline{12.035} & 12.250 & 12.880 & 13.160 & 12.909 & 12.780 & 14.718  \\ 
& MASE& \textbf{\textcolor{black}{1.612}} & \textcolor{black}{1.731} & 1.808 & 1.631 & 2.095 & \underline{1.623} & 1.625 & 1.698 & 1.836 & 1.775 & 1.771 & 1.756 & 2.408  \\ 
& OWA & \textbf{\textcolor{black}{0.857}} & \textcolor{black}{0.913} & 0.94 & 0.874 & 1.051 & \underline{0.869} & \underline{0.869} & 0.896 & 0.955 & 0.949 & 0.939 & 0.930 & 1.172  \\ 
\hline

\end{tabular}
\label{tab:short-term-forecast-full}
\end{table*}

\section{Full Few-shot Forecasting Results}\label{appendix:few-shot}

\begin{table*}[!h]
\caption{{Detailed few-shot learning results on 10\% training data}}
\setlength{\tabcolsep}{1.15pt}
\scriptsize
\centering

\begin{tabular}{cc|cc|cc|cc|cc|cc|cc|cc|cc|cc|cc|cc|cc}
\hline
\multicolumn{2}{c|}{Methods} &
  \multicolumn{2}{c|}{\textbf{$\mathbf{S^2}$IP-LLM}} &
  \multicolumn{2}{c|}{\textbf{Time-LLM(G)}}&
  \multicolumn{2}{c|}{\textbf{OFA}} &
  \multicolumn{2}{c|}{\textbf{iTrans.}} &
  \multicolumn{2}{c|}{\textbf{Dlinear}} &
  \multicolumn{2}{c|}{\textbf{PatchTST}} &
  \multicolumn{2}{c|}{\textbf{TimesNet}} &
  \multicolumn{2}{c|}{\textbf{FEDformer}} &
      \multicolumn{2}{c|}{\textbf{Autoformer}} &
  \multicolumn{2}{c|}{\textbf{Stationary}} &
  \multicolumn{2}{c|}{\textbf{ETSformer}} &
  \multicolumn{2}{c}{\textbf{LightTS}}   \\ \hline
\multicolumn{1}{c|}{$\text{Data.}$} &   \multicolumn{1}{c|}{$\text{Hori.}$}  &
  \multicolumn{1}{c}{MSE} &
  \multicolumn{1}{c|}{MAE} &
  \multicolumn{1}{c}{MSE} &
  \multicolumn{1}{c|}{MAE} &
  \multicolumn{1}{c}{MSE} &
  \multicolumn{1}{c|}{MAE} &
  \multicolumn{1}{c}{MSE} &
  \multicolumn{1}{c|}{MAE} &
  \multicolumn{1}{c}{MSE} &
  \multicolumn{1}{c|}{MAE} &
  \multicolumn{1}{c}{MSE} &
  \multicolumn{1}{c|}{MAE} &
  \multicolumn{1}{c}{MSE} &
  \multicolumn{1}{c|}{MAE} &
  \multicolumn{1}{c}{MSE} &
  \multicolumn{1}{c|}{MAE} &
  \multicolumn{1}{c}{MSE} &
  \multicolumn{1}{c|}{MAE} &
  \multicolumn{1}{c}{MSE} &
  \multicolumn{1}{c|}{MAE} &
  \multicolumn{1}{c}{MSE} &
  \multicolumn{1}{c|}{MAE} &
  \multicolumn{1}{c}{MSE} &
  \multicolumn{1}{c}{MAE} \\ \hline

\multicolumn{1}{l|}{\multirow{5}{*}{\rotatebox{90}{\textbf{Weather}}}}

& 96  & \textbf{{0.159}} & \textbf{{0.210}} & \underline{{0.160}} & \underline{{0.213}} & 0.163 & 0.215 & 0.253 & 0.307 & 0.171 & 0.224 & 0.165 & 0.215 & 0.184 & 0.230 & 0.188 & 0.253 & 0.221 & 0.297 & 0.192 & 0.234 & 0.199 & 0.272 & 0.217 & 0.269 \\
\multicolumn{1}{l|}{}                       
& 192 & \textbf{{0.200}} & \textbf{{0.251}} & \underline{{0.204}} & \underline{{0.254}} & 0.210 & \underline{0.254} & 0.292 & 0.328 & 0.215 & 0.263 & 0.210 & 0.257 & 0.245 & 0.283 & 0.250 & 0.304 & 0.270 & 0.322 & 0.269 & 0.295 & 0.279 & 0.332 & 0.259 & 0.304 \\
\multicolumn{1}{l|}{}                       
& 336 & {0.257} & {0.293} & \textbf{{0.255}} & \textbf{{0.291}} & \underline{0.256} & \underline{0.292} & 0.322 & 0.346 & 0.258 & 0.299 & 0.259 & 0.297 & 0.305 & 0.321 & 0.312 & 0.346 & 0.320 & 0.351 & 0.370 & 0.357 & 0.356 & 0.386 & 0.303 & 0.334  \\ 
\multicolumn{1}{l|}{}                      
& 720 & \textbf{{0.317}} & \textbf{{0.335}} & {0.329} & {0.345} & 0.321 & \underline{0.339} & 0.365 & 0.374 & \underline{0.320} & 0.346 & 0.332 & 0.346 & 0.381 & 0.371 & 0.387 & 0.393 & 0.390 & 0.396 & 0.441 & 0.405 & 0.437 & 0.448 & 0.377 & 0.382 \\
\multicolumn{1}{l|}{}                       
& Avg & \textbf{{0.233}} & \textbf{{0.272}} & \underline{{0.237}} & \underline{{0.275}} & 0.238 & \underline{0.275} & 0.308 & 0.338 & 0.241 & 0.283 & 0.242 & 0.279 & 0.279 & 0.301 & 0.284 & 0.324 & 0.300 & 0.342 & 0.318 & 0.323 & 0.318 & 0.360 & 0.289 & 0.322  \\

\hline

\multicolumn{1}{l|}{\multirow{5}{*}{\rotatebox{90}{\textbf{Electricity}}}} 

& 96  & {0.143} & {0.243} & \textbf{{0.137}} & \underline{{0.240}} &  \underline{0.139} &  \textbf{0.237} & 0.154 & 0.257 & 0.150 & 0.253 & 0.140 & 0.238 & 0.299 & 0.373 & 0.231 & 0.323 & 0.261 & 0.348 & 0.420 & 0.466 & 0.599 & 0.587 & 0.350 & 0.425\\
\multicolumn{1}{l|}{}                       
& 192 & \underline{{0.159}} & {0.258} & \underline{{0.159}} & {0.258} & \textbf{0.156} & \textbf{0.252} & 0.171 & 0.272 & 0.164 & 0.264 & 0.160 & \underline{0.255} & 0.305 & 0.379 & 0.261 & 0.356 & 0.338 & 0.406 & 0.411 & 0.459 & 0.620 & 0.598 & 0.376 & 0.448 \\
\multicolumn{1}{l|}{}                       
& 336 & \textbf{{0.170}} & \textbf{{0.269}} & {0.181} & {0.278} & \underline{0.175} & \underline{0.270} & 0.196 & 0.295  & 0.181 & 0.282 & 0.180 & 0.276 & 0.319 & 0.391 & 0.360 & 0.445 & 0.410 & 0.474 & 0.434 & 0.473 & 0.662 & 0.619 & 0.428 & 0.485 \\ 
\multicolumn{1}{l|}{}                      
& 720 & \underline{{0.230}} & \textbf{{0.315}} & {0.232} & \underline{{0.317}} & 0.233 & \underline{0.317} & 0.263  & 0.348  & \textbf{0.223} & 0.321 & 0.241 & 0.323 & 0.369 & 0.426 & 0.530 & 0.585 & 0.715 & 0.685 & 0.510 & 0.521 & 0.757 & 0.664 & 0.611 & 0.597 \\
\multicolumn{1}{l|}{}                       
& Avg & \textbf{{0.175}} & \underline{{0.271}} & {0.177} & {0.273} & \underline{0.176} & \textbf{0.269} & 0.196  &  0.293 &    0.180 & 0.280 & 0.180 & 0.273 & 0.323 & 0.392 & 0.346 & 0.427 & 0.431 & 0.478 & 0.444 & 0.480 & 0.660 & 0.617 & 0.441 & 0.489 \\

\hline

\multicolumn{1}{l|}{\multirow{5}{*}{\rotatebox{90}{\textbf{Traffic}}}}

& 96  & \textbf{{0.403}} & \underline{{0.293}} & \underline{{0.406}} & {0.295} & 0.414 & 0.297 &  0.448 & 0.329  & 0.419 & 0.298 & \textbf{0.403} & \textbf{0.289} & 0.719 & 0.416 & 0.639 & 0.400 & 0.672 & 0.405 & 1.412 & 0.802 & 1.643 & 0.855 & 1.157 & 0.636\\
\multicolumn{1}{l|}{}                       
& 192 & \textbf{{0.412}} & \textbf{{0.295}} & {0.416} & {0.300} & 0.426 & 0.301 & 0.487  & 0.360  & 0.434 & 0.305 & \underline{0.415} & \underline{0.296} & 0.748 & 0.428 & 0.637 & 0.416 & 0.727 & 0.424 & 1.419 & 0.806 & 1.641 & 0.854 & 1.207 & 0.661\\
\multicolumn{1}{l|}{}                       
& 336 & \underline{{0.427}} & {0.316} & {0.430} & {0.309} & 0.434 & \textbf{0.303} &  0.514 &  0.372  & 0.449 & 0.313 & \textbf{0.426} & \underline{0.304} & 0.853 & 0.471 & 0.655 & 0.427 & 0.749 & 0.454 & 1.443 & 0.815 & 1.711 & 0.878 & 1.334 & 0.713\\ 
\multicolumn{1}{l|}{}                      
& 720 &  \underline{{0.469}} & \underline{{0.325}} & \textbf{{0.467}} & \textbf{{0.324}} & 0.487 & 0.337 & 0.532  &  0.383  & 0.484 & 0.336 & 0.474 & 0.331 & 1.485 & 0.825 & 0.722 & 0.456 & 0.847 & 0.499 & 1.539 & 0.837 & 2.660 & 1.157 & 1.292 & 0.726\\
\multicolumn{1}{l|}{}                       
& Avg & \textbf{{0.427}} & \underline{{0.307}} & \underline{{0.429}} & \underline{{0.307}} & 0.440 & 0.310 &  0.495 &  0.361 & 0.447 & 0.313 & 0.430 & \textbf{0.305} & 0.951 & 0.535 & 0.663 & 0.425 & 0.749 & 0.446 & 1.453 & 0.815 & 1.914 & 0.936 & 1.248 & 0.684  \\

\hline

\multicolumn{1}{l|}{\multirow{5}{*}{\rotatebox{90}{\textbf{ETTh1}}}}

& 96 & \underline{{0.481}} & \underline{{0.474}} & 0.720 & 0.533 & \textbf{0.458} & \textbf{0.456} & 0.790 & 0.586 & 0.492 & 0.495 & 0.516 & 0.485 & 0.861 & 0.628 & 0.512 & 0.499 & 0.613 & 0.552 & 0.918 & 0.639 & 1.112 & 0.806 & 1.298 & 0.838 \\
\multicolumn{1}{l|}{}                       
& 192 & \textbf{{0.518}} & \textbf{{0.491}} & 0.747 & 0.545 & 0.570 & \underline{0.516} & 0.837 & 0.609 & \underline{0.565} & 0.538 & 0.598 & 0.524 & 0.797 & 0.593 & 0.624 & 0.555 & 0.722 & 0.598 & 0.915 & 0.629 & 1.155 & 0.823 & 1.322 & 0.854 \\
\multicolumn{1}{l|}{}                       
& 336 & {0.664} & {0.570} & 0.793 & 0.551 & \textbf{0.608} & \textbf{0.535} & 0.780 & 0.575 & 0.721 & 0.622 & \underline{0.657} & \underline{0.550} & 0.941 & 0.648 & 0.691 & 0.574 & 0.750 & 0.619 & 0.939 & 0.644 & 1.179 & 0.832 & 1.347 & 0.870  \\
\multicolumn{1}{l|}{}                      
& 720 & \textbf{{0.711}} & \textbf{{0.584}} & 0.880 & \textbf{0.584} & \underline{0.725} & \underline{0.591} & 1.234 & 0.811 & 0.986 & 0.743 & 0.762 & 0.610 & 0.877 & 0.641 & 0.728 & 0.614 & 0.721 & 0.616 & 0.887 & 0.645 & 1.273 & 0.874 & 1.534 & 0.947   \\
\multicolumn{1}{l|}{}                       
& Avg & \underline{{0.593}} & \underline{{0.529}} & 0.785 & 0.553 & \textbf{0.590} & \textbf{0.525} & 0.910 & 0.860  & 0.691 & 0.600 & 0.633 & 0.542 & 0.869 & 0.628 & 0.639 & 0.561 & 0.702 & 0.596 & 0.915 & 0.639 & 1.180 & 0.834 & 1.375 & 0.877 \\ 

\hline

\multicolumn{1}{l|}{\multirow{5}{*}{\rotatebox{90}{\textbf{ETTh2}}}}

& 96 & {0.354} & {0.400} & \underline{{0.334}} & \underline{{0.381}} & \textbf{0.331} & \textbf{0.374} & 0.404 & 0.435 & 0.357 & 0.411 & 0.353 & 0.389 & 0.378 & 0.409 & 0.382 & 0.416 & 0.413 & 0.451 & 0.389 & 0.411 & 0.678 & 0.619 & 2.022 & 1.006 \\
\multicolumn{1}{l|}{}                       
& 192 & \textbf{{0.401}} & {0.423} & {0.430} & {0.438} & \underline{0.402} & \textbf{0.411} & 0.470 & 0.474 & 0.569 & 0.519 & 0.403 & \underline{0.414} & 0.490 & 0.467 & 0.478 & 0.474 & 0.474 & 0.477 & 0.473 & 0.455 & 0.785 & 0.666 & 2.329 & 1.104  \\
\multicolumn{1}{l|}{}                       
& 336 & {0.442} & {0.450} & {0.449} & {0.458} & \textbf{0.406} & \textbf{0.433} & 0.489 & 0.485 & 0.671 & 0.572 & \underline{0.426} & \underline{0.441} & 0.537 & 0.494 & 0.504 & 0.501 & 0.547 & 0.543 & 0.477 & 0.472 & 0.839 & 0.694 & 2.453 & 1.122 \\ 
\multicolumn{1}{l|}{}                      
& 720 & {0.480} & {0.486}  & {0.485} & {0.490} & \textbf{0.449} & \textbf{0.464} & 0.593 & 0.538 & 0.824 & 0.648 & \underline{0.477} & \underline{0.480} & 0.510 & 0.491 & 0.499 & 0.509 & 0.516 & 0.523 & 0.507 & 0.480 & 1.273 & 0.874 & 3.816 & 1.407  \\
\multicolumn{1}{l|}{}                       
& Avg & {0.419} & {0.439}  & {0.424} & {0.441} & \textbf{0.397} & \textbf{0.421} & 0.489 & 0.483  & 0.605 & 0.538 & \underline{0.415} & \underline{0.431} & 0.479 & 0.465 & 0.466 & 0.475 & 0.488 & 0.499 & 0.462 & 0.455 & 0.894 & 0.713 & 2.655 & 1.160  \\ 
\hline
\multicolumn{1}{l|}{\multirow{5}{*}{\rotatebox{90}{\textbf{ETTm1}}}}

& 96  & \underline{{0.388}} & \underline{{0.401}} & {0.412} & {0.422} & 0.390 & 0.404  & 0.709 & 0.556 & \textbf{0.352} & \textbf{0.392} & 0.410 & 0.419 & 0.583 & 0.501 & 0.578 & 0.518 & 0.774 & 0.614 & 0.761 & 0.568 & 0.911 & 0.688 & 0.921 & 0.682 \\
\multicolumn{1}{l|}{}                       
& 192 & \underline{{0.422}} & \underline{{0.421}} & {0.447} & {0.438} & 0.429 & 0.423 & 0.717 & 0.548 & \textbf{0.382} & \textbf{0.412} & 0.437 & 0.434 & 0.630 & 0.528 & 0.617 & 0.546 & 0.754 & 0.592 & 0.781 & 0.574 & 0.955 & 0.703 & 0.957 & 0.701 \\
\multicolumn{1}{l|}{}                       
& 336 & \underline{{0.456}} & \textbf{{0.430}} & {0.497} & {0.465} & 0.469 & 0.439 & 0.735 & 0.575 & \textbf{0.419} & \underline{0.434} & 0.476 & 0.454 & 0.725 & 0.568 & 0.998 & 0.775 & 0.869 & 0.677 & 0.803 & 0.587 & 0.991 & 0.719 & 0.998 & 0.716   \\
\multicolumn{1}{l|}{}                      
& 720 & \underline{{0.554}} & \underline{{0.490}} & {0.594} & {0.521} & 0.569 & 0.498 & 0.752 & 0.584 & \textbf{0.490} & \textbf{0.477} & 0.681 & 0.556 & 0.769 & 0.549 & 0.693 & 0.579 & 0.810 & 0.630 & 0.844 & 0.581 & 1.062 & 0.747 & 1.007 & 0.719\\
\multicolumn{1}{l|}{}                       
& Avg & \underline{{0.455}} & \underline{{0.435}}  & {0.487} & {0.461} & 0.464 & 0.441 & 0.728 & 0.565 & \textbf{0.411} & \textbf{0.429} & 0.501 & 0.466 & 0.677 & 0.537 & 0.722 & 0.605 & 0.802 & 0.628 & 0.797 & 0.578 & 0.980 & 0.714 & 0.971 & 0.705 \\ 
\hline

\multicolumn{1}{l|}{\multirow{5}{*}{\rotatebox{90}{\textbf{ETTm2}}}} 

& 96 & {0.192} & \underline{{0.274}} & {0.224} & {0.296} & \textbf{0.188} & \textbf{0.269} & 0.245 & 0.322 & 0.213 & 0.303 & \underline{0.191} & \underline{0.274} & 0.212 & 0.285 & 0.291 & 0.399 & 0.352 & 0.454 & 0.229 & 0.308 & 0.331 & 0.430 & 0.813 & 0.688 \\
\multicolumn{1}{l|}{}                       
& 192 & \textbf{{0.246}} & \underline{{0.313}} & {0.260} & {0.317} & \underline{0.251} & \textbf{0.309} & 0.274 & 0.338 & 0.278 & 0.345 & 0.252 & 0.317 & 0.270 & 0.323 & 0.307 & 0.379 & 0.694 & 0.691 & 0.291 & 0.343 & 0.400 & 0.464 & 1.008 & 0.768 \\
\multicolumn{1}{l|}{}                       
& 336 & \textbf{{0.301}} & \textbf{{0.340}} & {0.312} & \underline{{0.349}} & 0.307 & 0.346 & 0.361 & 0.394 & 0.338 & 0.385 & \underline{0.306} & 0.353 & 0.323 & 0.353 & 0.543 & 0.559 & 2.408 & 1.407 & 0.348 & 0.376 & 0.469 & 0.498 & 1.031 & 0.775  \\ 
\multicolumn{1}{l|}{}                      
& 720 & \textbf{{0.400}} & \textbf{{0.403}} & \underline{{0.424}} & \underline{{0.416}} & 0.426 & 0.417 & 0.467 & 0.442 & 0.436 & 0.440 & 0.433 & 0.427 & 0.474 & 0.449 & 0.712 & 0.614 & 1.913 & 1.166 & 0.461 & 0.438 & 0.589 & 0.557 & 1.096 & 0.791 \\
\multicolumn{1}{l|}{}                       
& Avg & \textbf{{0.284}} & \textbf{{0.332}} & {0.305} & {0.344} & \underline{0.293} & \underline{0.335} & 0.336 & 0.373 & 0.316 & 0.368 & 0.296 & 0.343 & 0.320 & 0.353 & 0.463 & 0.488 & 1.342 & 0.930 & 0.332 & 0.366 & 0.447 & 0.487 & 0.987 & 0.756  \\

\bottomrule

\end{tabular}
\label{tab:long-term-forecast_few10}
\end{table*}

\begin{table*}[!h]
\caption{{Detailed few-shot learning results on 5\% training data.'-' means 5\% data is not sufficient to constitute a training set.}}
\setlength{\tabcolsep}{1.15pt}
\scriptsize
\centering

\begin{tabular}{cc|cc|cc|cc|cc|cc|cc|cc|cc|cc|cc|cc|cc}
\hline
\multicolumn{2}{c|}{Methods} &
  \multicolumn{2}{c|}{\textbf{$\mathbf{S^2}$IP-LLM}} &
  \multicolumn{2}{c|}{\scriptsize\textbf{Time-LLM(G)}}&
  \multicolumn{2}{c|}{\textbf{OFA}} &
  \multicolumn{2}{c|}{\textbf{iTrans.}} &
  \multicolumn{2}{c|}{\textbf{Dlinear}} &
  \multicolumn{2}{c|}{\textbf{PatchTST}} &
  \multicolumn{2}{c|}{\textbf{TimesNet}} &
  \multicolumn{2}{c|}{\textbf{FEDformer}} &
  \multicolumn{2}{c|}{\textbf{Autoformer}} &
  \multicolumn{2}{c|}{\textbf{Stationary}} &
  \multicolumn{2}{c|}{\textbf{ETSformer}} &
  \multicolumn{2}{c}{\textbf{LightTS}}  \\ \hline
\multicolumn{1}{c|}{$\text{Data.}$} &   \multicolumn{1}{c|}{$\text{Hori.}$} &
  \multicolumn{1}{c}{MSE} &
  \multicolumn{1}{c|}{MAE} &
  \multicolumn{1}{c}{MSE} &
  \multicolumn{1}{c|}{MAE} &
  \multicolumn{1}{c}{MSE} &
  \multicolumn{1}{c|}{MAE} &
  \multicolumn{1}{c}{MSE} &
  \multicolumn{1}{c|}{MAE} &
  \multicolumn{1}{c}{MSE} &
  \multicolumn{1}{c|}{MAE} &
  \multicolumn{1}{c}{MSE} &
  \multicolumn{1}{c|}{MAE} &
  \multicolumn{1}{c}{MSE} &
  \multicolumn{1}{c|}{MAE} &
  \multicolumn{1}{c}{MSE} &
  \multicolumn{1}{c|}{MAE} &
  \multicolumn{1}{c}{MSE} &
  \multicolumn{1}{c|}{MAE} &
  \multicolumn{1}{c}{MSE} &
  \multicolumn{1}{c|}{MAE} &
  \multicolumn{1}{c}{MSE} &
  \multicolumn{1}{c|}{MAE} &
  \multicolumn{1}{c}{MSE} &
  \multicolumn{1}{c}{MAE} \\ \hline

\multicolumn{1}{c|}{\multirow{5}{*}{\rotatebox{90}{\textbf{Weather}}}} 

& 96  & \underline{\textcolor{black}{0.175}} & \underline{\textcolor{black}{0.228}} & \textcolor{black}{0.176} & \textcolor{black}{0.230} & \underline{0.175} & 0.230 & 0.264 & 0.307 & 0.184 & 0.242 & \textbf{0.171} & \textbf{0.224} & 0.207 & 0.253 & 0.229 & 0.309 & 0.227 & 0.299 & 0.215 & 0.252 & 0.218 & 0.295 & 0.230 & 0.285 \\
\multicolumn{1}{l|}{}                       
& 192 & \textbf{\textcolor{black}{0.225}} & \textbf{\textcolor{black}{0.271}} & \underline{\textcolor{black}{0.226}} & \underline{\textcolor{black}{0.275}} & 0.227 & 0.276 & 0.284 & 0.326 & 0.228 & 0.283 & 0.230 & 0.277 & 0.272 & 0.307 & 0.265 & 0.317 & 0.278 & 0.333 & 0.290 & 0.307 & 0.294 & 0.331 & 0.274 & 0.323 \\
\multicolumn{1}{l|}{}                       
& 336 & \underline{0.282} & \textbf{0.321} & \textcolor{black}{0.292} & \textcolor{black}{0.325} & 0.286 & 0.322 & 0.323 & 0.349  & \textbf{0.279} & \underline{0.322} & 0.294 & 0.326 & 0.313 & 0.328 & 0.353 & 0.392 & 0.351 & 0.393 & 0.353 & 0.348 & 0.359 & 0.398 & 0.318 & 0.355  \\ 
\multicolumn{1}{l|}{}                      
& 720 & \textbf{0.361} & \textbf{0.371} & \underline{\textcolor{black}{0.364}} & \underline{\textcolor{black}{0.375}} & 0.366 & 0.379 & 0.366 & 0.375 & 0.364 & 0.388 & 0.384 & 0.387 & 0.400 & 0.385 & 0.391 & 0.394 & 0.387 & 0.389 & 0.452 & 0.407 & 0.461 & 0.461 & 0.401 & 0.418 \\
\multicolumn{1}{l|}{}                       
& Avg & \textbf{0.260} & \textbf{0.297} & 0.264 & \underline{0.301} & \underline{0.263} & \underline{0.301} & 0.309 &  0.339 & 0.263 & 0.308 & 0.269 & 0.303 & 0.298 & 0.318 & 0.309 & 0.353 & 0.310 & 0.353 & 0.327 & 0.328 & 0.333 & 0.371 & 0.305 & 0.345  \\

\hline

\multicolumn{1}{c|}{\multirow{5}{*}{\rotatebox{90}{\textbf{Electricity}}}} 

& 96  & \textcolor{black}{0.148} & \textcolor{black}{0.248} & \textcolor{black}{0.148} & \textcolor{black}{0.248} & \textbf{0.143} & \textbf{0.241} & 0.162 & 0.264 & 0.150 & 0.251 & \underline{0.145} & \underline{0.244} & 0.315 & 0.389 & 0.235 & 0.322 & 0.297 & 0.367 & 0.484 & 0.518 & 0.697 & 0.638 & 0.639 & 0.609\\
\multicolumn{1}{l|}{}                       
& 192 & \textbf{0.159} & \textbf{0.255} & \underline{0.160} & \underline{0.257} & \textbf{0.159} & \textbf{0.255} & 0.180 & 0.278 & 0.163 & 0.263 & 0.163 & 0.260 & 0.318 & 0.396 & 0.247 & 0.341 & 0.308 & 0.375 & 0.501 & 0.531 & 0.718 & 0.648 & 0.772 & 0.678\\
\multicolumn{1}{l|}{}                       
& 336 & \textbf{0.175} & \textbf{0.271}  & \textcolor{black}{0.183} & \textcolor{black}{0.282} & \underline{0.179} & \underline{0.274} & 0.207 & 0.305 & \textbf{0.175} & 0.278 & 0.183 & 0.281 & 0.340 & 0.415 & 0.267 & 0.356 & 0.354 & 0.411 & 0.574 & 0.578 & 0.758 & 0.667 & 0.901 & 0.745\\ 
\multicolumn{1}{l|}{}                      
& 720 & 0.235 & 0.326 & \textcolor{black}{0.236} & \textcolor{black}{0.329} & \underline{0.233} & \underline{0.323} & 0.258 & 0.339 & \textbf{0.219} & \textbf{0.311} & 0.233 & 0.323 & 0.635 & 0.613 & 0.318 & 0.394 & 0.426 & 0.466 & 0.952 & 0.786 & 1.028 & 0.788 & 1.200 & 0.871\\
\multicolumn{1}{l|}{}                       
& Avg & 0.179 & 0.275 & 0.181 & 0.279 & \underline{0.178} & \textbf{0.273} & 0.201 & 0.296 & \textbf{0.176} & \underline{0.275} & 0.181 & 0.277 & 0.402 & 0.453 & 0.266 & 0.353 & 0.346 & 0.404 & 0.627 & 0.603 & 0.800 & 0.685 & 0.878 & 0.725  \\

\hline

\multicolumn{1}{c|}{\multirow{5}{*}{\rotatebox{90}{\textbf{Traffic}}}} 

& 96  & \underline{0.410} & \underline{0.288} & \textcolor{black}{0.414} & \textcolor{black}{0.293} & 0.419 & 0.298 & 0.431 & 0.312 & 0.427 & 0.304 & \textbf{0.404} & \textbf{0.286} & 0.854 & 0.492 & 0.670 & 0.421 & 0.795 & 0.481 & 1.468 & 0.821 & 1.643 & 0.855 & 1.157 & 0.636\\
\multicolumn{1}{l|}{}                       
& 192 & \underline{0.416} & \underline{0.298} & \textcolor{black}{0.419} & \textcolor{black}{0.300} & 0.434 & 0.305 & 0.456 & 0.326 & 0.447 & 0.315 & \textbf{0.412} & \textbf{0.294} & 0.894 & 0.517 & 0.653 & 0.405 & 0.837 & 0.503 & 1.509 & 0.838 & 1.856 & 0.928 & 1.688 & 0.848\\
\multicolumn{1}{c|}{}                       
& 336 & \textbf{0.435} & \underline{0.313} & \underline{\textcolor{black}{0.438}} & \textcolor{black}{0.315} & 0.449 & \underline{0.313} & 0.465 & 0.334 & 0.478 & 0.333 & 0.439 & \textbf{0.310} & 0.853 & 0.471 & 0.707 & 0.445 & 0.867 & 0.523 & 1.602 & 0.860 & 2.080 & 0.999 & 1.826 & 0.903\\ 
\multicolumn{1}{l|}{}                      
& 720 & - & - & - & - & - & - & - & - &- & - & - & - & - & - &  - & - & - & - & - & - &- & -& - & -  \\
\multicolumn{1}{l|}{}                       
& Avg & \underline{0.420} & \underline{0.299} & 0.423 & 0.302 & 0.434 & 0.305 & 0.450 & 0.324 & 0.450 & 0.317 & \textbf{0.418} & \textbf{0.296} & 0.867 & 0.493 & 0.676 & 0.423 & 0.833 & 0.502 & 1.526 & 0.839 & 1.859 & 0.927 & 1.557 & 0.795   \\

\hline

\multicolumn{1}{c|}{\multirow{5}{*}{\rotatebox{90}{\textbf{ETTh1}}}}

& 96 & \textbf{\textcolor{black}{0.500}} & \textbf{\textcolor{black}{0.493}} & 0.732 & 0.556 & \underline{0.543}  & 0.506 & 0.808 & 0.610 & 0.547 & \underline{0.503} & 0.557 & 0.519 & 0.892 & 0.625 & 0.593 & 0.529 & 0.681 & 0.570 & 0.952 & 0.650 &1.169 & 0.832& 1.483 & 0.910 \\
\multicolumn{1}{c|}{}                       
& 192 & \underline{\textcolor{black}{0.690}} & \textbf{\textcolor{black}{0.539}} & 0.872 & 0.604 & 0.748 & 0.580  & 0.928 & 0.658 & 0.720 & 0.604 & 0.711 & 0.570 & 0.940 & 0.665 & \textbf{0.652} & \underline{0.563} & 0.725 & 0.602 & 0.943 & 0.645 &1.221 & 0.853 &  1.525 & 0.930 \\
\multicolumn{1}{c|}{}                       
& 336 & \textcolor{black}{0.761} & \textcolor{black}{0.620} & 1.071 & 0.721 & \underline{0.754} & \underline{0.595} & 1.475 & 0.861 & 0.984 & 0.727 & 0.816 & 0.619  & 0.945 & 0.653 & \textbf{0.731} & \textbf{0.594} & 0.761 & 0.624 &  0.935 & 0.644  & 1.179 & 0.832 & 1.347 & 0.870  \\
\multicolumn{1}{c|}{}                      
& 720 & - & - & - & - & - & - & - & - &- & - & - & - & - & - &  - & - & - & - & - & - &- & -& - & -   \\
\multicolumn{1}{c|}{}                       
& Avg & \textbf{\textcolor{black}{0.650}} & \textbf{\textcolor{black}{0.550}} & 0.891 & 0.627 & 0.681 & \underline{0.560} & 1.070 & 0.710 & 0.750 & 0.611 & 0.694 & 0.569 & 0.925 & 0.647 & \underline{0.658} & 0.562 & 0.722 & 0.598 & 0.943 & 0.646 & 1.189 &  0.839 & 1.451 & 0.903 \\ 

\hline

\multicolumn{1}{c|}{\multirow{5}{*}{\rotatebox{90}{\textbf{ETTh2}}}} 

& 96 & \textbf{\textcolor{black}{0.363}} & \textbf{\textcolor{black}{0.409}} & \textcolor{black}{0.399} & \underline{\textcolor{black}{0.420}} & \underline{0.376} & 0.421 & 0.397 & 0.427 & 0.442 & 0.456 & 0.401 & 0.421 & 0.409 & 0.420 & 0.390 & 0.424 & 0.428 & 0.468 & 0.408 & 0.423 & 0.678 & 0.619 & 2.022 & 1.006  \\
\multicolumn{1}{c|}{}                       
& 192 & \textbf{\textcolor{black}{0.375}} & \textbf{\textcolor{black}{0.411}} & 0.487 & 0.479 & \underline{0.418} & \underline{0.441}& 0.438 & 0.445 & 0.617 & 0.542 & 0.452 & 0.455 & 0.483 & 0.464 & 0.457 & 0.465 & 0.496 & 0.504 & 0.497 & 0.468 & 0.845 & 0.697 & 3.534 & 1.348  \\
\multicolumn{1}{c|}{}                       
& 336 & \textbf{\textcolor{black}{0.403}} & \textbf{\textcolor{black}{0.421}} & 0.858 & 0.660 & \underline{0.408} & \underline{0.439} & 0.631 & 0.553 & 1.424 & 0.849 & 0.464 & 0.469 & 0.499 & 0.479 & 0.477 & 0.483 & 0.486 & 0.496 & 0.507 & 0.481 & 0.905 & 0.727 & 4.063 & 1.451 \\
\multicolumn{1}{c|}{}                      
& 720 & - & - & -& - & - & - & - & - &- & - & - & - & - & - &  - & - & - & - & - & - &- & -& - & -  \\
\multicolumn{1}{c|}{}                       
& Avg & \textbf{\textcolor{black}{0.380}} & \textbf{\textcolor{black}{0.413}} & 0.581 & 0.519 & \underline{0.400} & \underline{0.433} & 0.488 & 0.475 & 0.694 & 0.577 & 0.827 & 0.615 & 0.439 & 0.448 & 0.463 & 0.454 & 0.441 & 0.457 & 0.470 & 0.489 & 0.809 & 0.681 & 3.206 & 1.268  \\ 
\hline
\multicolumn{1}{c|}{\multirow{5}{*}{\rotatebox{90}{\textbf{ETTm1}}}}

& 96  & \underline{\textcolor{black}{0.357}} & \underline{\textcolor{black}{0.390}} & \textcolor{black}{0.422} & \textcolor{black}{0.424} & 0.386 & 0.405 & 0.589 & 0.510 & \textbf{0.332} & \textbf{0.374} & 0.399 & 0.414 & 0.606 & 0.518 & 0.628 & 0.544 & 0.726 & 0.578 & 0.823 & 0.587 & 1.031 & 0.747 & 1.048 & 0.733 \\
\multicolumn{1}{c|}{}                       
& 192 & \underline{\textcolor{black}{0.432}} & \underline{\textcolor{black}{0.434}} & \textcolor{black}{0.448} & \textcolor{black}{0.440} & 0.440 & 0.438 & 0.703 & 0.565 & \textbf{0.358} & \textbf{0.390} & 0.441 & 0.436 & 0.681 & 0.539 & 0.666 & 0.566 & 0.750 & 0.591 & 0.844 & 0.591 & 1.087 & 0.766 & 1.097 & 0.756 \\
\multicolumn{1}{c|}{}                       
& 336 & \underline{\textcolor{black}{0.440}} & \underline{\textcolor{black}{0.442}} & \textcolor{black}{0.519} & \textcolor{black}{0.482} & 0.485 & 0.459 & 0.898 & 0.641 & \textbf{0.402} & \textbf{0.416} & 0.499 & 0.467 & 0.786 & 0.597 & 0.807 & 0.628 & 0.851 & 0.659 & 0.870 & 0.603 & 1.138 & 0.787 & 1.147 & 0.775  \\
\multicolumn{1}{c|}{}                      
& 720 & 0.593 & 0.521 & 0.708 & 0.573 & \underline{0.577} & \underline{0.499} & 0.948 & 0.671 & \textbf{0.511} & \textbf{0.489} & 0.767 & 0.587 & 0.796 & 0.593 & 0.822 & 0.633 & 0.857 & 0.655 & 0.893 & 0.611 & 1.245 & 0.831 & 1.200 & 0.799\\
\multicolumn{1}{c|}{}                       
& Avg & \underline{0.455} & \underline{0.446} & 0.524 & 0.479 & 0.472 & 0.450 & 0.784 & 0.596 & \textbf{0.400} & \textbf{0.417} & 0.526 & 0.476 & 0.717 & 0.561 & 0.730 & 0.592 & 0.796 & 0.620 & 0.857 & 0.598 & 1.125 & 0.782 & 1.123 & 0.765 \\ 
\hline

\multicolumn{1}{c|}{\multirow{5}{*}{\rotatebox{90}{\textbf{ETTm2}}}}

& 96  & \textbf{\textcolor{black}{0.197}} & \textbf{\textcolor{black}{0.278}} & \textcolor{black}{0.225} & \textcolor{black}{0.300} & \underline{0.199} & \underline{0.280} &  0.265 & 0.339 & 0.236 & 0.326 & 0.206 & 0.288 & 0.220 & 0.299 & 0.229 & 0.320 & 0.232 & 0.322 & 0.238 & 0.316 & 0.404 & 0.485 & 1.108 & 0.772 \\
\multicolumn{1}{c|}{}                       
& 192 & \textbf{\textcolor{black}{0.254}} & \underline{\textcolor{black}{0.322}} & \textcolor{black}{0.275} & \textcolor{black}{0.334}& \underline{0.256} & \textbf{0.316} & 0.310 & 0.362 & 0.306 & 0.373 & 0.264 & 0.324 & 0.311 & 0.361 & 0.394 & 0.361 & 0.291 & 0.357 & 0.298 & 0.349 & 0.479 & 0.521 & 1.317 & 0.850 \\
\multicolumn{1}{c|}{}                       
& 336 & \textbf{\textcolor{black}{0.315}} & \textbf{\textcolor{black}{0.350}} & \textcolor{black}{0.339} & \textcolor{black}{0.371} & \underline{0.318} & \underline{0.353} & 0.373 & 0.399 & 0.380 & 0.423 & 0.334 & 0.367 & 0.338 & 0.366 & 0.378 & 0.427 & 0.478 & 0.517 & 0.353 & 0.380 & 0.552 & 0.555 & 1.415 & 0.879  \\ 
\multicolumn{1}{c|}{}                      
& 720 & \textbf{\textcolor{black}{0.421}} & \textbf{\textcolor{black}{0.421}} & \textcolor{black}{0.464} & \textcolor{black}{0.441} & 0.460 & 0.436 & 0.478 & 0.454 & 0.674 & 0.583 & \underline{0.454} & \underline{0.432} & 0.509 & 0.465 & 0.523 & 0.510 & 0.553 & 0.538 & 0.475 & 0.445 & 0.701 & 0.627 & 1.822 & 0.984 \\
\multicolumn{1}{c|}{}                       
& Avg & \textbf{\textcolor{black}{0.296}} & \textbf{\textcolor{black}{0.342}} & \textcolor{black}{0.325} & \textcolor{black}{0.361} & \underline{0.308} & \underline{0.346} & 0.356 & 0.388 & 0.399 & 0.426 & 0.314 & 0.352 & 0.344 & 0.372 & 0.381 & 0.404 & 0.388 & 0.433 & 0.341 & 0.372 & 0.534 & 0.547 & 1.415 & 0.871  \\

\bottomrule

\end{tabular}
\label{tab:long-term-forecast_few5}
\end{table*}

Table~\ref{tab:long-term-forecast_few10} and Table~\ref{tab:long-term-forecast_few5} show the full few-short forecasting experiment results with 10\% and 5\% of the training data respectively.

\section{Ablation Studies and Parameter Sensitivity}\label{Appendix: ablation}

\begin{figure}[ht]
\begin{center}
\centerline{\includegraphics[width=1\columnwidth]{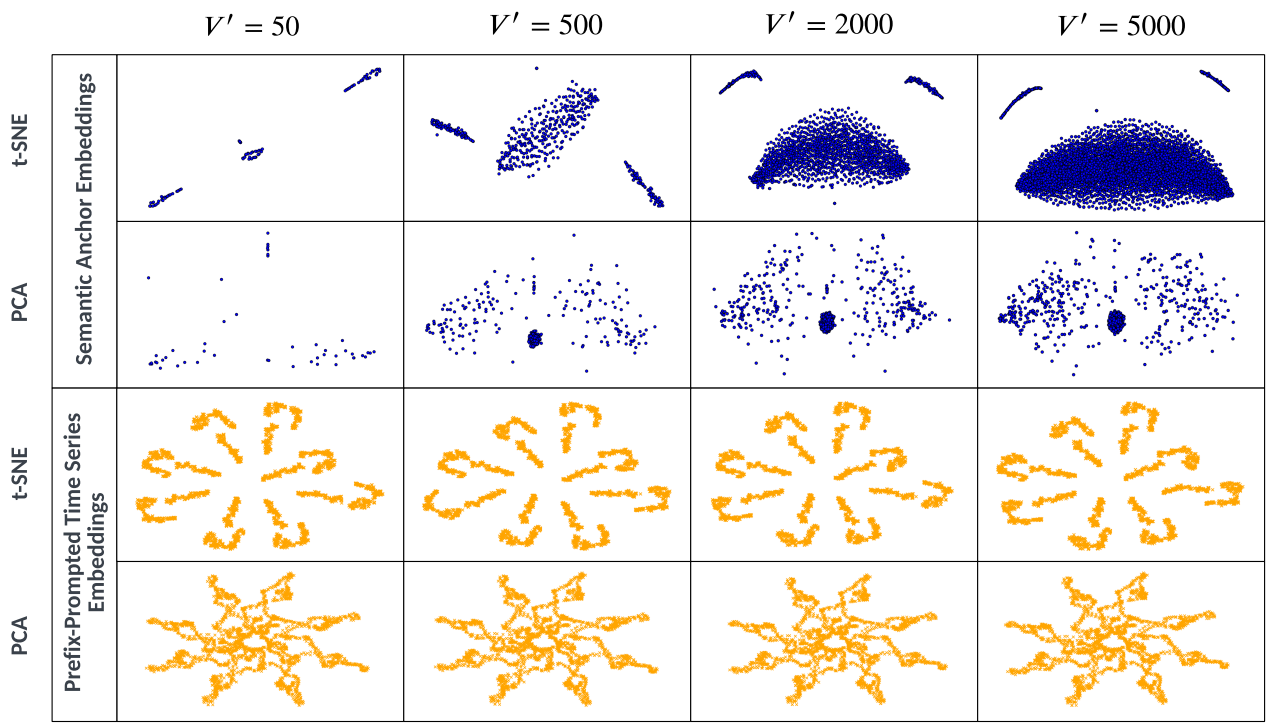}}
\caption{The t-SNE and PCA plots of semantic anchor and prefix-prompted time series embeddings with different $V'$} 
\label{tsne-vis_3}
\end{center}
\end{figure}

We provide the t-SNE and PCA visualization of semantic anchor and prefix-prompted time series embeddings with different $V'$ in Figure~\ref{tsne-vis_3}. We observe semantic anchor embeddings display a continued spanning pattern among the joint space, whereas the prompted time series representation shows only a slight visual difference. It is reasonable since it is primarily controlled by the scaling factor $\lambda$.

\textcolor{black}{}

\section{Visualization}\label{greedy}

\begin{figure}[!h] 
  \centering
  \includegraphics[scale=0.35]{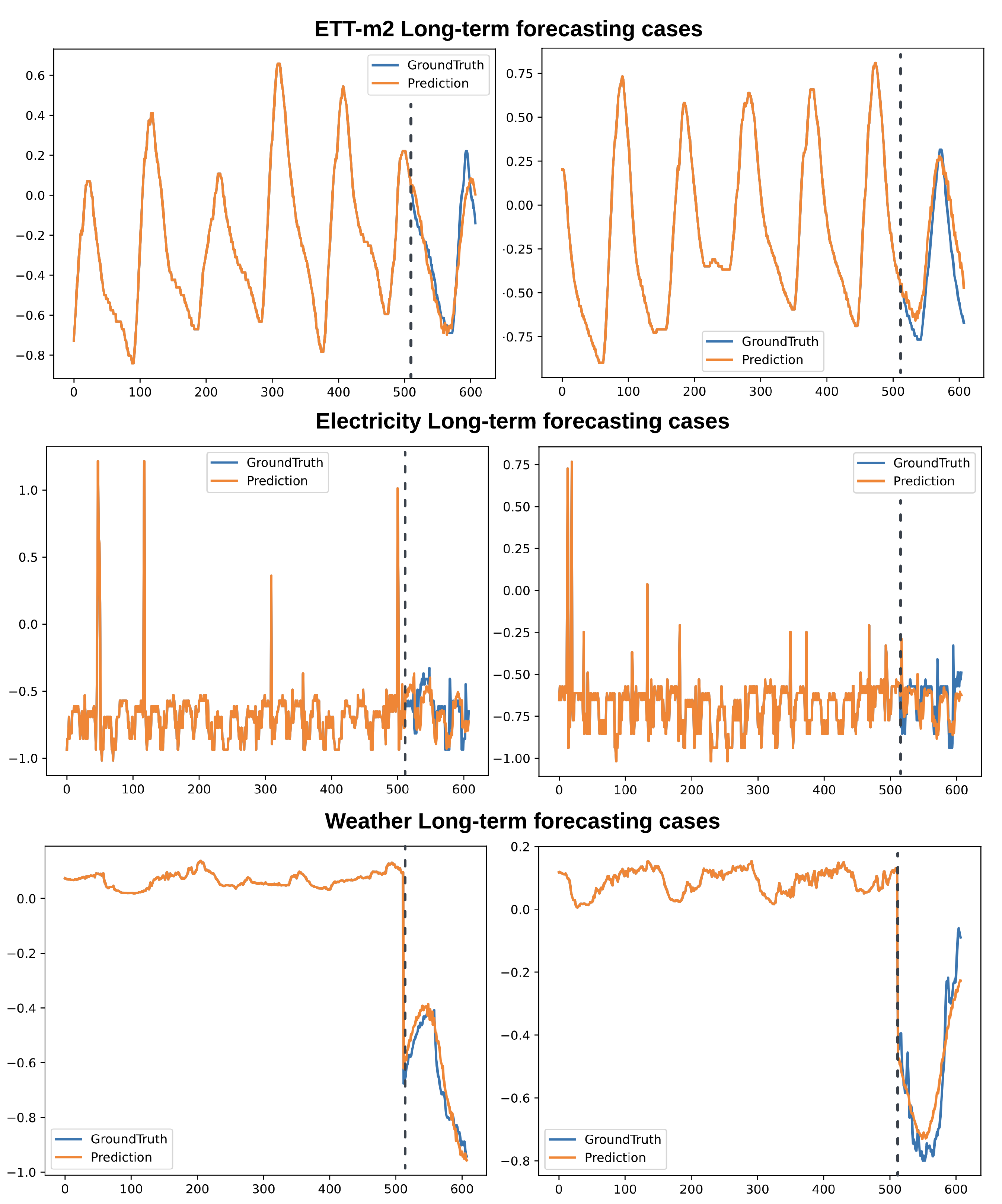} %
  \caption{Long-term forecasting visualization cases for ETTm2, Electricity, and Weather. \textcolor{blue}{Blue} lines are the ground truths and \textcolor{orange}{orange} lines are the model predictions. The vertical line indicates where the prediction starts.}
  \label{vis} 
\end{figure}

In this section, we provide the visualizations of the forecasting cases of $S^2$IP-LLM on ETTm2, Electricity, and Weather datasets under the input-512-predict-96 setting. As shown in Figure \ref{vis}, $S^2$IP-LLM achieves exceptionally good forecasting results across various datasets.

\begin{table*}[t]
\centering
\small
\caption{Ablation studies on ETTh2 and ETTm2 in predicting 96 and 192 steps (MSE reported).}
\begin{tabular}{c|cccc}
\hline\hline
\multicolumn{1}{c|}{\multirow{2}{*}{\begin{tabular}[c]{@{}c@{}}Ablation \\ Setting\end{tabular}}} & \multicolumn{4}{c}{Long-term Forecasting}                                                                                   \\ \cline{2-5} 
\multicolumn{1}{c|}{}                                                                             & \multicolumn{1}{c}{ETTh2-96} & \multicolumn{1}{c}{ETTh2-192} & \multicolumn{1}{c}{ETTm2-96} & \multicolumn{1}{c}{ETTm2-192} \\ \hline
\multicolumn{1}{c|}{w/o  Prompt \& Alignment and w/o Decomposition}& 

0.289  & 0.358 & 0.170  & 0.231  \\

w/ Prompt \& Alignment and w/o Decomposition & 0.287 & 0.353 & 0.166 & 0.228 \\

w/ Prompt \& Alignment and w/ Decomposition & 0.278 & 0.346 & 0.165 & 0.222 \\

\hline

\end{tabular}
\label{Ablation}
\end{table*}

\end{document}